\documentclass{article}
\usepackage{ijcai21}

\usepackage{times}
\usepackage{soul}
\usepackage{url}
\usepackage[hidelinks]{hyperref}
\usepackage[utf8]{inputenc}
\usepackage[small]{caption}
\usepackage{graphicx}
\usepackage{subcaption}
\usepackage{amsmath}
\usepackage{amsthm}
\usepackage{booktabs}
\usepackage{algorithm}
\usepackage{algorithmic}
\usepackage{comment}
\DeclareMathOperator*{\argmin}{arg\,min}

\newtheorem*{problem}{Problem}

\author{
Yan Zhou$^1$
\and
Murat Kantarcioglu$^1$\and
Chris Clifton$^{2}$
\affiliations
$^1$Department of Computer Science, University of Texas at Dallas, Richardson, TX 75080\\
$^2$ Department of Computer Sciences, Purdue University,  West Lafayette, Indiana, 47907-2107
\emails
\{yan.zhou2, muratk\}@utdallas.edu,
clifton@cs.purdue.edu
}
\title{Improving Fairness of AI Systems with Lossless De-biasing}

\begin{document}

\maketitle

\begin{abstract}
In today's society, AI systems are increasingly used to make critical decisions  such as credit scoring and patient triage. However, great convenience brought by AI systems comes with troubling prevalence of bias against underrepresented groups. Mitigating bias in AI systems to increase overall fairness has emerged as an important challenge. Existing studies on mitigating bias in AI systems focus on eliminating sensitive demographic information embedded in data. Given the temporal and contextual complexity of conceptualizing fairness, lossy treatment of demographic information may contribute to an unnecessary trade-off between accuracy and fairness, especially when demographic attributes and class labels are correlated. In this paper,  we present an information-lossless de-biasing technique that targets the scarcity of data in the disadvantaged group. Unlike the existing work, we demonstrate, both theoretically and empirically, that oversampling underrepresented groups can not only mitigate algorithmic bias in AI systems that consistently predict a favorable outcome for a certain group, but improve overall accuracy by  mitigating class imbalance within data that leads to a bias towards the majority class.  
%MK: Let us make this more strong. can we add things like we are the first ones to do this and that?
%Yan: addressed by emphasizing we can mitigate bias and improve accuracy.
%MK: Can we also discuss the improvements compared to existign techniques and why this is a better approach.
%Yan: ours improve both fairness and accuracy in theory
We demonstrate the effectiveness of our technique on real datasets using a variety of fairness metrics.
\end{abstract}

\section{Introduction}
\label{sec:intro}
Algorithmic bias has become a known issue as AI algorithms are criticized for reflecting and potentially exacerbating human biases in data. 
Existence of bias in automated systems for job hiring, credit lending, health care, predictive policing, and criminal sentencing inevitably perpetuates inequalities in the society. 
%MK: Adding an example for these biases, please check.
For example, the controversy of the well known COMPAS recidivism risk assessment tool~\cite{COMPAS} reveals that black defendants are disproportionately labeled  as more likely to re-offend in the future.  
De-biasing techniques requisite to keep the algorithms fair progress naturally in three directions: pre-processing, in-processing, and post-processing. Pre-processing is concerned with manipulating data to mitigate bias before they are used for training~\cite{10.1145/2783258.2783311,10.5555/3042817.3042973,NIPS2017_9a49a25d,10.1007/s10115-011-0463-8}. In-processing tackles bias at the algorithmic level, modifying algorithms to remove bias~\cite{pmlr-v80-madras18a,9aa5ba8a091248d597ff7cf0173da151,10.1145/3278721.3278779,pmlr-v80-kearns18a,10.1145/3287560.3287592,10.1145/3287560.3287586,10.5555/3120007.3120011}, while post-processing is re-adjusting decision output to reflect justification weighted on fairness~\cite{10.5555/3295222.3295319,NIPS2016_9d268236,10.1109/ICDM.2012.45,kim2019multiaccuracy}. 
%MK: Maybe add few examples of the proposed fairness criterias 
Meanwhile, many fairness metrics have been proposed~\cite{pmlr-v28-zemel13,9101635}, implying the interlocking complexity of the fairness problem in automated AI systems. 

Existing pre-processing de-biasing techniques focus  on transforming a given input by editing its features and labels, assigning weights to selected training samples, or learning a latent representation excluding sensitive features to increase fairness of the trained model. The general consensus is that the cause of algorithmic bias lies within the inbuilt biases in data, favoring the privileged group as decisions are made. This interpretation of bias is imperfect, but it  simplifies and allows for the  formalization of shaping fairness into AI models.   Transforming data to mitigate bias has an obvious disadvantage of information loss. In this paper, we investigate the feasibility and effectiveness of generating synthetic data to augment the representation of the unprivileged demographic group and eliminate the inherent bias in data, and furthermore eliminate base rate difference in favored predictions by trained models. An important challenge is that when inserting synthetic data through oversampling the underrepresented region in the input, we should not risk losing or altering the information in the original input while improving fairness.

The main contributions of this paper include:
\begin{itemize}
\itemsep=0pt
\item An information-lossless de-biasing technique that generates synthetic data to augment the underrepresented region of the input;
\item Theoretical justification of the proposed de-biasing technique;
%MK: Can we add a sentence showing how our technique does better compared to existing techniques.
\item Extensive empirical studies that compare our de-biasing technique with the existing pre-processing, in-processing, and post-processing techniques using a variety of fairness metrics.
%\item empirical investigation 
\end{itemize}

The rest of the paper is organized as follows: Section~\ref{sec:related} discusses existing related work. Section~\ref{sec:method} presents our synthetic data generation de-biasing technique and Section~\ref{sec:exp} presents our experimental results. Section~\ref{sec:conclude} concludes our work and discusses future directions. 

\section{Related Work}
\label{sec:related}

De-biasing through pre-processing typically transforms training data by reducing the influence of demographic changes on the positive base rate. Feldman et~al.~\shortcite{10.1145/2783258.2783311} propose to alter the unprotected attributes to remove disparate impact. Zemel et~al.~\shortcite{10.5555/3042817.3042973} encode the input with a latent representation that obfuscates sensitive attributes. Optimized preprocessing~\cite{NIPS2017_9a49a25d} learns a probabilistic transformation for the input to improve group fairness while limiting individual data distortion. Reweighing~\cite{10.1007/s10115-011-0463-8} assigns different weights to selected samples to ensure fairer predictions by trained classifiers. 

In-processing de-biasing is done at the algorithmic level where learning algorithms are tweaked to ensure fairness. Some de-biasing techniques coupled with adversarial objectives consider learning fair representation under the constraint of different adversarial objectives for group fairness~\cite{pmlr-v80-madras18a,9aa5ba8a091248d597ff7cf0173da151}. De-biasing with adversarial learning~\cite{10.1145/3278721.3278779} aims to maximize predictive accuracy while minimizing adversary's ability to predict sensitive attributes. Other in-processing de-biasing techniques focus on fairness constraints on structured subgroups~\cite{pmlr-v80-kearns18a,10.1145/3287560.3287592}, training an optimized classifier with respect to a given fairness metric ~\cite{10.1145/3287560.3287586}, or adding a regularization term to the  objective against discrimination~\cite{10.5555/3120007.3120011}.

Post-processing techniques modify output labels to meet different fairness objectives. Some calibrates classifier outputs to ensure equalized odds~\cite{10.5555/3295222.3295319,NIPS2016_9d268236}, some makes favorable predictions for unprivileged groups and unfavorable predictions for privileged groups in the vicinity of decision boundaries~\cite{10.1109/ICDM.2012.45}, and in the case where only black-box access is granted, a classifier satisfying multi-accuracy fairness conditions can be learned to improve fairness and subgroup accuracy~\cite{kim2019multiaccuracy}.

%MK: Can we add couple of sentences on why our technique is better and different then existing alternatives?
Our de-biasing technique is in line with the philosophy of de-biasing through pre-processing. Our technique is unique in that we create datasets that not only address historical bias, but also boost the representation of minority groups, i.e. correcting bias introduced by the algorithms in a way that is algorithm-agnostic. 

\section{De-biasing with Synthetic Data}
\label{sec:method}

In this section, we first motivate the idea of de-biasing AI systems by oversampling the underrepresented region in the input. Next, we provide theoretical justification and demonstrate an empirical bound of predictive difference between the privileged group and the unprivileged group. 

\subsection{Lossless De-biasing}
\label{sec:oversample}

Existing de-biasing techniques tend to carry ``hostile'' attitudes to the demographic attributes that are often linked to the socio-economic status of an individual and may potentially cause bias. As a result, the majority of the efforts are focused on transforming the input data or surgically modifying AI algorithms so that they become agnostic to demographic information. This type of treatment is inevitably lossy, and sometimes unnecessarily trading accuracy for fairness. 
%MK: Please check the next sentence modification.
More importantly, many such techniques tend to hard code fairness criteria (e.g., equalized odds criteria)  while in reality it is rarely the case that different parties reach a consensus on how fairness gets defined. Leaving room for disagreement in de-biasing techniques is equally critical in order to avoid bias in fairness conceptualization. 

Instead of reducing the demographic information naturally represented in a given input, we reverse the course of completely eliminating the entanglement between demographic attributes, fairness, and social cost, expanding the favorable decision region for the underrepresented group while preserving the utility of the given demographic information. 
Since algorithmic bias is often rooted in the lack of favorable representation of the disadvantaged group in training data, a straightforward approach to promoting fairness is to oversample favorable representation of this  group of individuals who are similarly situated except for their disadvantaged socio-economic status. 
%MK: Please give an example here to make the idea clear. I added an example check.
For example, we can generate samples of black defendants that do not re-offend in the COMPAS data collection.
By generating synthetic data that favorably represents the disadvantaged group, we can, hypothetically, correct flaws in the training data that either inherits historical inequalities or systematically skews the decision towards a favored group. In a later section, we provide theoretical justification for this idea. 

We now formally define the problem of de-biasing an AI algorithm that predicts either favorable or unfavorable binary outcome for a given input. 
%MK: I wonder this formalisation is ok. Basically, this says we make two distributions similar. I believe we are just giving more examples, so that AI model can learn accurate patterns. In other words, our goal is not to make blacks looks like white but make sure that we have enough black defendants who do not reoffend so that the model can learn reasonable patterns from these examples. I guess given the time let us keep this definition for now and formalization. 
\begin{problem}
Given a domain $\mathcal{X}$ with a probability distribution $D$ over $\mathcal{X}$ and a target function $f: \mathcal{X} \rightarrow [0,1]$, let $D_p$ and $D_u$ be the distributions over the privileged group and the unprivileged group in $\mathcal{X}$, we formalize de-biasing as bounding the discrepancy between the distributions $D_p$ and $D_u'$ over the populations of the privileged group and the augmented unprivileged group given an arbitrary $\epsilon > 0$:
\[
 d(D_p, D_u') = 2 \sup_{s \in \mathcal{S}}|Pr_{D_p}(s) - Pr_{D_u'}(s)| \le \epsilon
\]  
where $D_u'$ is the distribution of the augmented unprivileged group and $\mathcal{S}$ is a collection of measurable subsets under $D_p$ and $D_u'$.  
\end{problem}
We formalize the problem of de-biasing as bounding variation divergence between distributions over the privileged group and the augmented unprivileged group. This formalization is inspired by the concept of transfer learning: if two populations only differ in demographic and socioeconomic background, models trained on one population (e.g. white defendants) should be readily applicable to the other (e.g. black defendants), with bounded predictive errors and positive base rate difference. Thus, our de-biasing technique is designed to augment data representing the unprivileged group so that the variation divergence for the two distributions $D_p$ and $D_u'$ is $\epsilon$-{\em close} with respect to $\mathcal{S}$ for an arbitrary $\epsilon > 0$. 

In the next section, we explain how to bound the difference in predictive errors and the discrepancy of favorable predictions between privileged and unprivileged groups using variation divergence,  and how the objective of de-biasing is related to bounding variation divergence between two distributions that can be estimated using samples from $D_p$ and $D_u'$.

\subsection{Bounding Difference in Predictive Errors}
\label{sec:bounderror}

Given a hypothesis $h$ trained on data from domain $\mathcal{X}$, the difference of predictive errors on data from $D_p$ and $D_u$ by $h$ is bounded with respect to the variation divergence between $D_p$ and $D_u$~\cite{10.5555/1316689.1316707}:  
\[
\epsilon_{D_p} (h) - \epsilon_{D_u} (h) \le  d(D_p, D_u) + \lambda
\]
where $\lambda = \argmin_{h \in \mathcal{H}} [\epsilon_{D_p} (h) + \epsilon_{D_u} (h)]$, which is the the combined error of the ideal joint hypothesis.

Notice the above bound is defined over variation divergence $d(D_p, D_u)$. Variation divergence cannot be accurately estimated from limited samples~\cite{Fortnow00testingthat}. Ben-David et al. proposed the $\mathcal{H}$-{\em divergence} to make feasible measuring divergence between two distributions $D$ and $D'$ over domain $\mathcal{X}$~\cite{36364}:
\begin{align*}
d_{\mathcal{H}}(D, D') & =  2 \sup_{h \in \mathcal{H}}[Pr_{D}[I(h)] - Pr_{D'}[I(h)]] \\
& \ge  2 |Pr_{D}[h(x)=1] - Pr_{D'}[h(x)=1]| 
\end{align*}
where $\mathcal{H}$ is a hypothesis class of finite VC dimension on $\mathcal{X}$, $I(h)$ is the set such that $\mathbf{x} \in I(h) \Leftrightarrow h(\mathbf{x})=1$.  Therefore, the discrepancy between favorable predictions by $h$ on data from $D_p$ and $D_u$ is bounded by the $\mathcal{H}$-{\em divergence} between $D_p$ and $D_u$:
\[
|Pr_{D}[h(x)=1] - Pr_{D'}[h(x)=1]| \le \frac{1}{2}d_{\mathcal{H}}(D_p, D_u).
\]

Kifer et al.~\shortcite{10.5555/1316689.1316707} provide a theoretical bound for the true $\mathcal{H}$-{\em divergence} given any $\delta \in (0,1)$ with probability at least $1-\delta$~\cite{36364}:
\[
d_{\mathcal{H}}(D, D') \le \hat{d}_{\mathcal{H}}(U, U') + 4  \sqrt{\frac{d \log(2m)+\log(\frac{2}{\delta})}{m} }
\]
where $U$ and $U'$ are samples of size $m$ from $D$ and $D'$. As the sample size increases, the empirical $\mathcal{H}$-{\em divergence} asymptotically approaches the true $\mathcal{H}$-{\em divergence}. 

With the concept of $\mathcal{H}$-{\em divergence}, we can bound the difference of predictive errors and the discrepancy of favorable predictions on data $U_p$ and $U_u$ of size $m$ from $D_p$ and $D_u$ as follows:
\begin{equation}
\label{eq:bias}
\Delta \epsilon(h(x)=1)  \le \frac{1}{2}\hat{d}_{\mathcal{H}}(U_p, U_u) + 2  \sqrt{\frac{d \log(2m)+\log(\frac{2}{\delta})}{m} } 
\end{equation}
\begin{equation}
\label{eq:error}
\Delta \epsilon(h) \le \frac{1}{2}\hat{d}_{\mathcal{H}}(U_p, U_u) + 2  \sqrt{\frac{d \log(2m)+\log(\frac{2}{\delta})}{m} } + \lambda
\end{equation}
where $\Delta \epsilon(h) = \epsilon_{D_p} (h) - \epsilon_{D_u} (h)$, $\Delta \epsilon(h(x)=1)=|Pr_{D}[h(x)=1] - Pr_{D'}[h(x)=1]|$, and $\lambda$ is the the combined error of the ideal joint hypothesis as defined earlier. 

As~(\ref{eq:bias}) and~(\ref{eq:error}) suggest, we can limit both the discrepancy of favorable predictions and the difference in predictive  errors by making the  distribution of the two groups $D_p$ and $D_u$ diverge less, especially when one group is underrepresented in terms of favorable prediction. %This is the incentive to oversample the underrepresented group that are treated less favorable by a learned hypothesis $h \in \mathcal{H}$ by generating synthetic data in that group.
In the next section, we discuss different scenarios of oversampling the underrepresented data to mitigate bias. 

\subsection{De-biasing Methods}
%MK: Please formally define what is majority, what is privileged and give example the context of crime justice system
Bias in an AI model may come from different sources. By oversampling the underrepresented population, we do not have to discern the source of bias, but instead consider scenarios where the majority or the minority of the demographic population is favored. Given a demographic attribute $A=\{a_1, a_2\}$ in a dataset $X$, the majority group defined on $A$ is $X_{A=a^*} \subset X$ if   $Pr(A=a^*|X) > Pr(A\neq a^*|X)$ where $a^*\in\{a_1,a_2\}$, and the privileged group defined on $A$ is $X_{A=a^*} \subset X$ if, historically, $Pr(y=1|X_{A=a^*}) > Pr(y=1|X_{A\neq a^*})$ where $a^*\in\{a_1,a_2\}$, $y \in \{0,1\}$ and $y=1$ is favored. For example, in the  COMPAS dataset, the black race is the majority and the white race is privileged. We investigate the following scenarios:
\begin{itemize}
\itemsep=0pt
\item[1.)] When the majority is privileged, that is, the majority group observes more favored predictions than the minority group, there are two different directions to proceed with oversampling as shown in Figure~\ref{fig:oversampling}:
\begin{itemize}
\item[a.)] We oversample by generating synthetic data in the underrepresented group with favored predictions;  
\item[b.)] We oversample in the privileged group with unfavored predictions.
\end{itemize}

\begin{figure}[!htb]
\centering
\begin{minipage}{.3\textwidth}
\centering
\includegraphics[width=\textwidth]{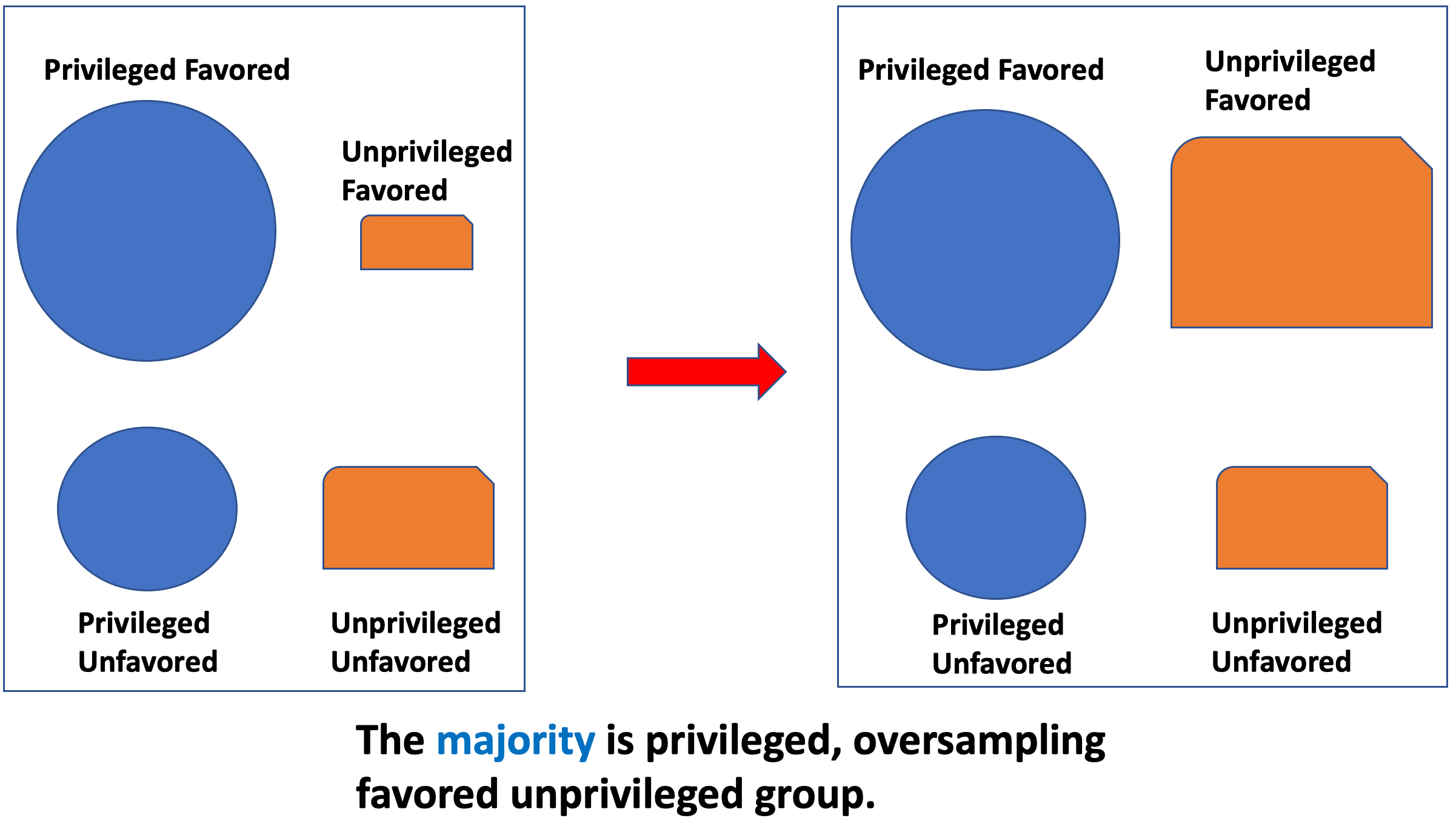}
\end{minipage}
\par\bigskip
\begin{minipage}{.3\textwidth}
\centering
\includegraphics[width=\textwidth]{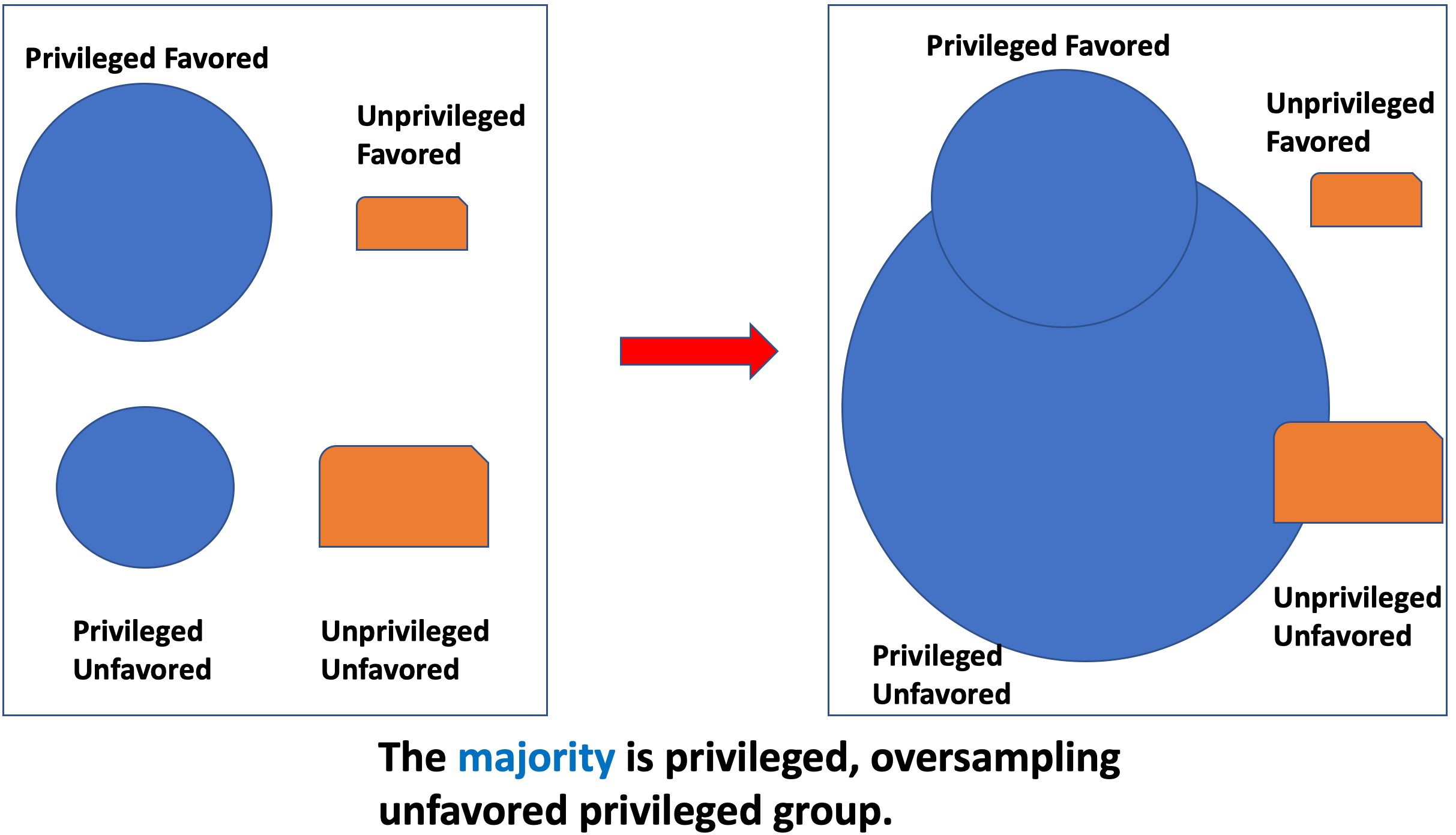}
\end{minipage}
\caption{\label{fig:oversampling} Oversampling when the majority of the population has more favored predictions.}
\end{figure}
Method 1(b) is proposed for comparison purpose, representing a possible use case of oversampling for de-biasing.  

\item[2.)] When the minority is privileged,  that is, the minority group observes more favored predictions, we oversample the minority group with unfavored predictions, as shown in Figure~\ref{fig:oversampling_minority}.
%MK: Please increase the fonts without making the figures bigger. I believe the fonts need to be increased for readability
% As it is they are not readable i meant, figure 1 and figure 2.
%Also figure two bottom left is too small and looks like there is a typo. Please update.
%Yan: fixed
\begin{figure}[!htb]
\centering
\begin{minipage}{.3\textwidth}
\centering
\includegraphics[width=\textwidth]{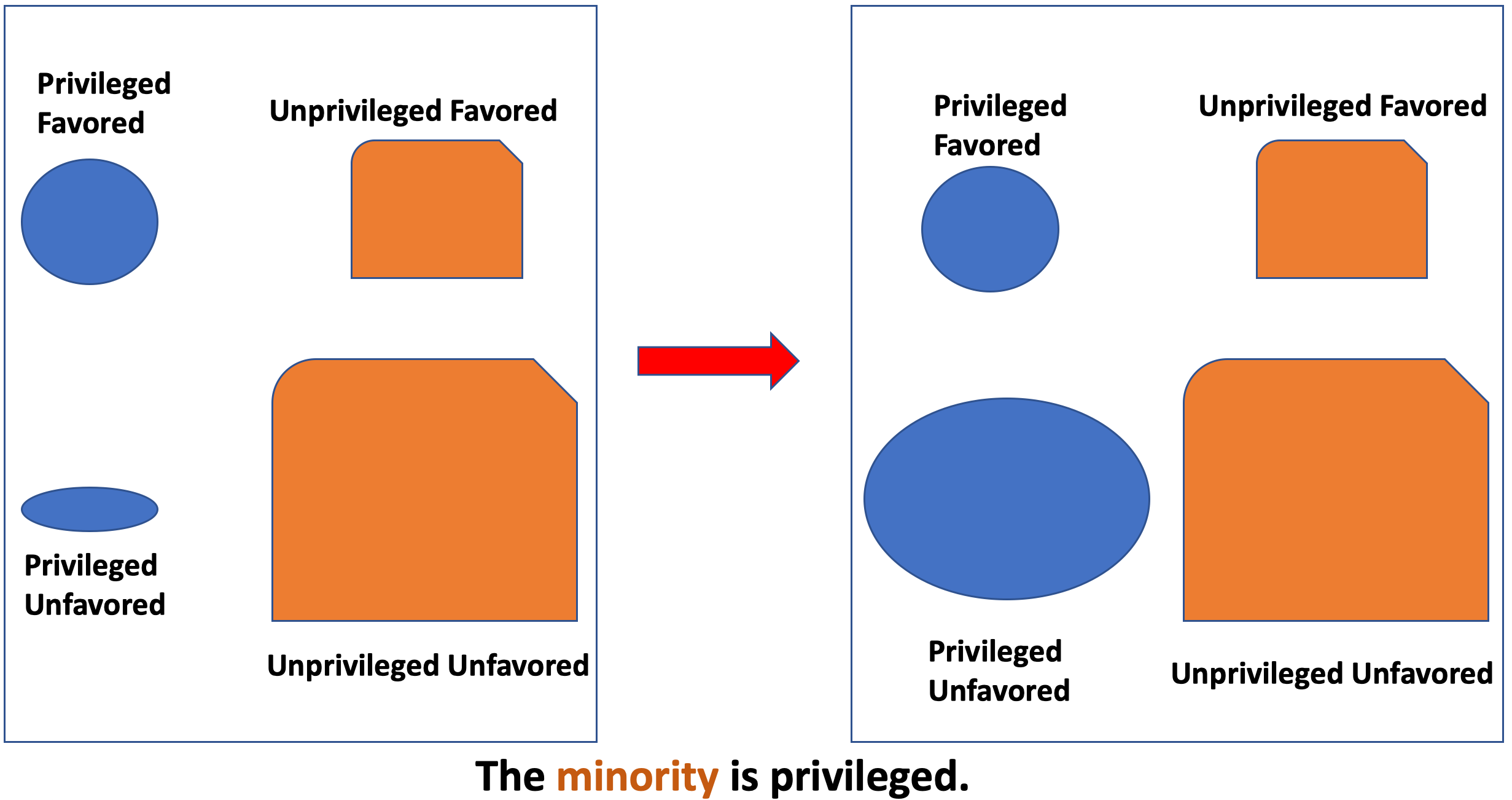}
\end{minipage}
\caption{\label{fig:oversampling_minority} Oversampling when the minority of the population has more favored predictions.}
\end{figure}
\end{itemize}
The idea of this de-biasing approach is to generate synthetic data to reduce the difference in positive base rate between the privileged and the unprivileged groups. When more underrepresented data is generated, the divergence between $D_p$ and $D_u$ is reduced, and consequently we bridge the gap between the two groups with fairer favorable predictions.

\section{Experimental Results}
\label{sec:exp}

We test our de-biasing technique on five data sets: {\em Adult}, {\em Compas}, {\em German Credit}, {\em Medical Expanse}, and {\em Bank} data~\cite{Dua:2019}. These datasets represent the two general cases where the majority is privileged and the minority is privileged. For the baseline  learning algorithms, we mainly use {\em Logistic Regression} (LR) and {\em Random Forest} (RF), and for de-biasing, we mainly compare our technique  to  {\em Reweighing} (pre-processing), {\em Prejudice Remover} (in-processing), and {\em Reject Option} (post-processing) de-biasing techniques. Additional experiments are also performed on the {\em Compas} data to investigate baseline algorithms SVM and Neural Network (NN), and other choices of mitigators such as {\em  Disparate Impact Remover}  (pre-processing), {\em Exponentiated Gradient Reduction} (in-processing), and {\em Calibrated EqOdds} (post-processing). 
We do not choose the {\em adversarial de-biasing} technique since it either fails to de-bias or suffers significant accuracy drop. We measure fairness using a number of individual and group fairness metrics, including {\em average odds difference}, {\em disparate impact}, {\em statistical parity difference}, {\em equal opportunity difference}, and {\em Theil index}. De-biasing algorithms used for comparison are implemented in the IBM AI Fairness 360 library~\cite{aif360-oct-2018}. Synthetic (non-existing) data is generated using SMOTE~\cite{smote_jair} until the other group is not disproportionately (dis)advantaged in training set.

\subsection{Adult Data}
\label{subsec:adult}
\begin{figure}[!htb]
\centering
\begin{subfigure}[b]{.225\textwidth}
\centering
\includegraphics[width=\textwidth]{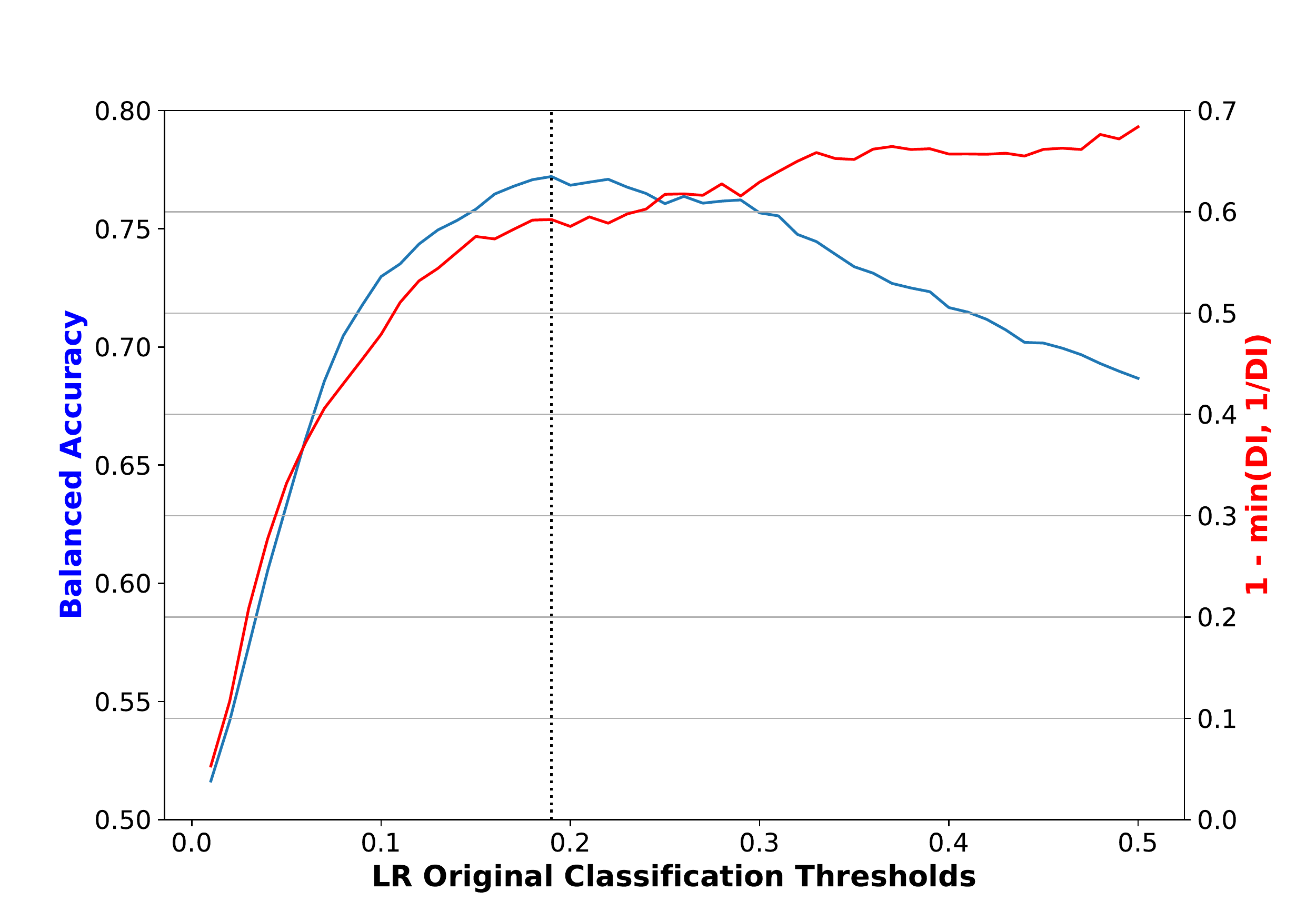}
\caption{\label{fig:LR_orig} Accuracy and Disparate Impact for Original LR.}
\end{subfigure}
\hfill
\begin{subfigure}[b]{.225\textwidth}
\centering
\includegraphics[width=\textwidth]{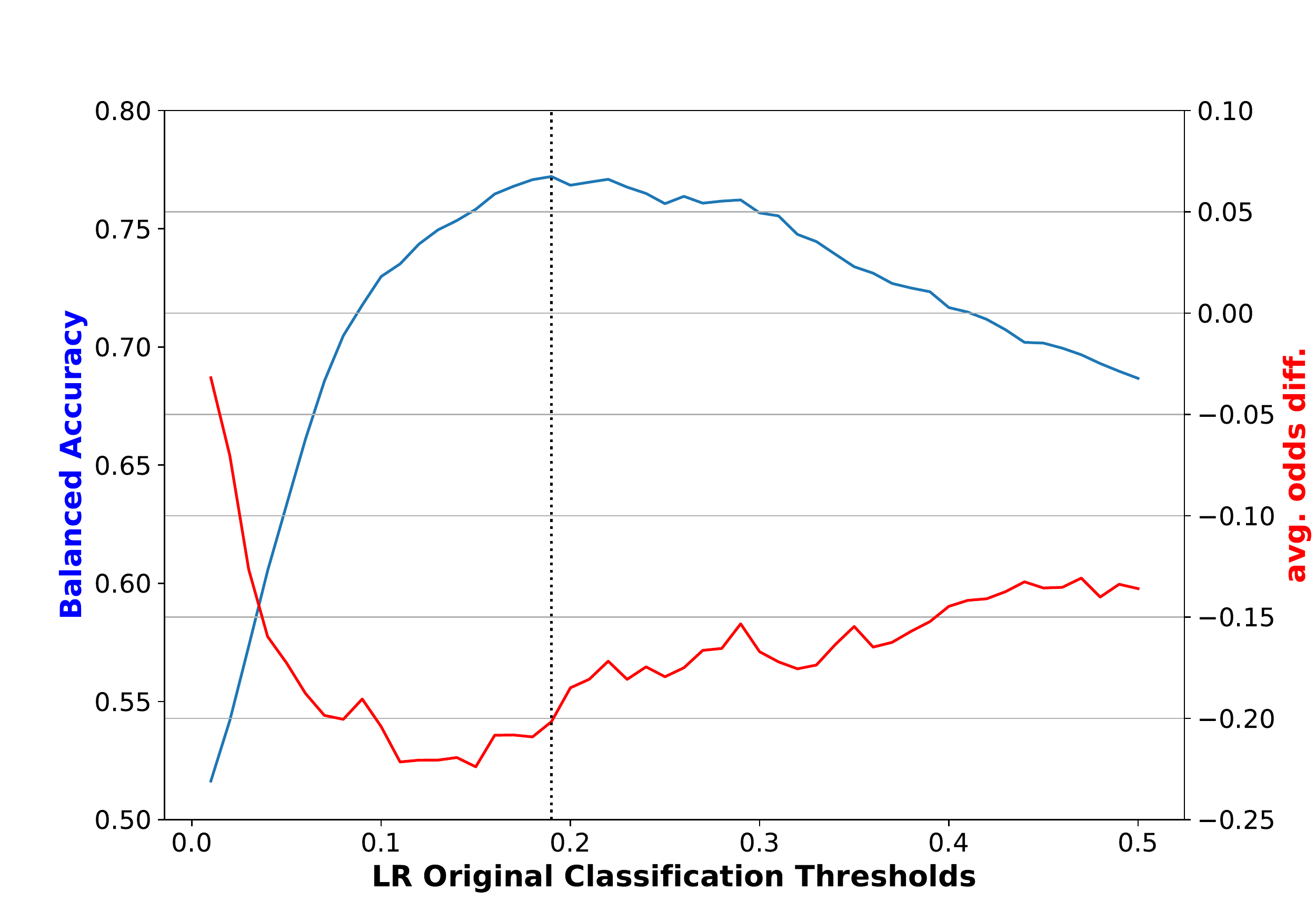}
\caption{\label{fig:LR_orig_odds} Accuracy and Avg. Odds Diff. for Original LR.}
\end{subfigure}
%\newline
%\end{figure}
%\begin{figure}[!htb]
%\centering
\begin{subfigure}[b]{.225\textwidth}
\centering
\includegraphics[width=\textwidth]{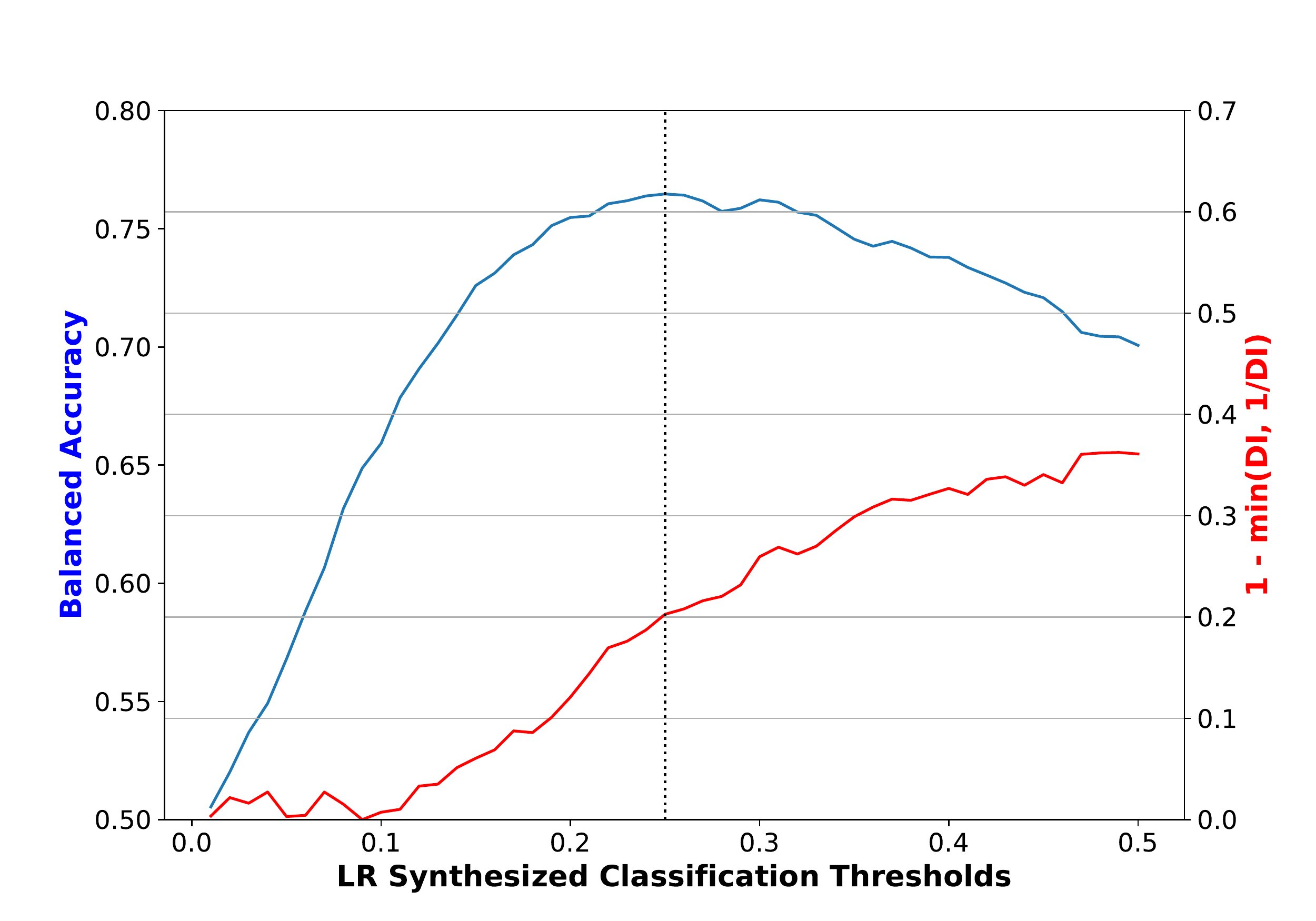}
\caption{\label{fig:LR_synth} Accuracy and Disparate Impact for LR with Synthetic Data.}
\end{subfigure}
\hfill
\begin{subfigure}[b]{.225\textwidth}
\centering
\includegraphics[width=\textwidth]{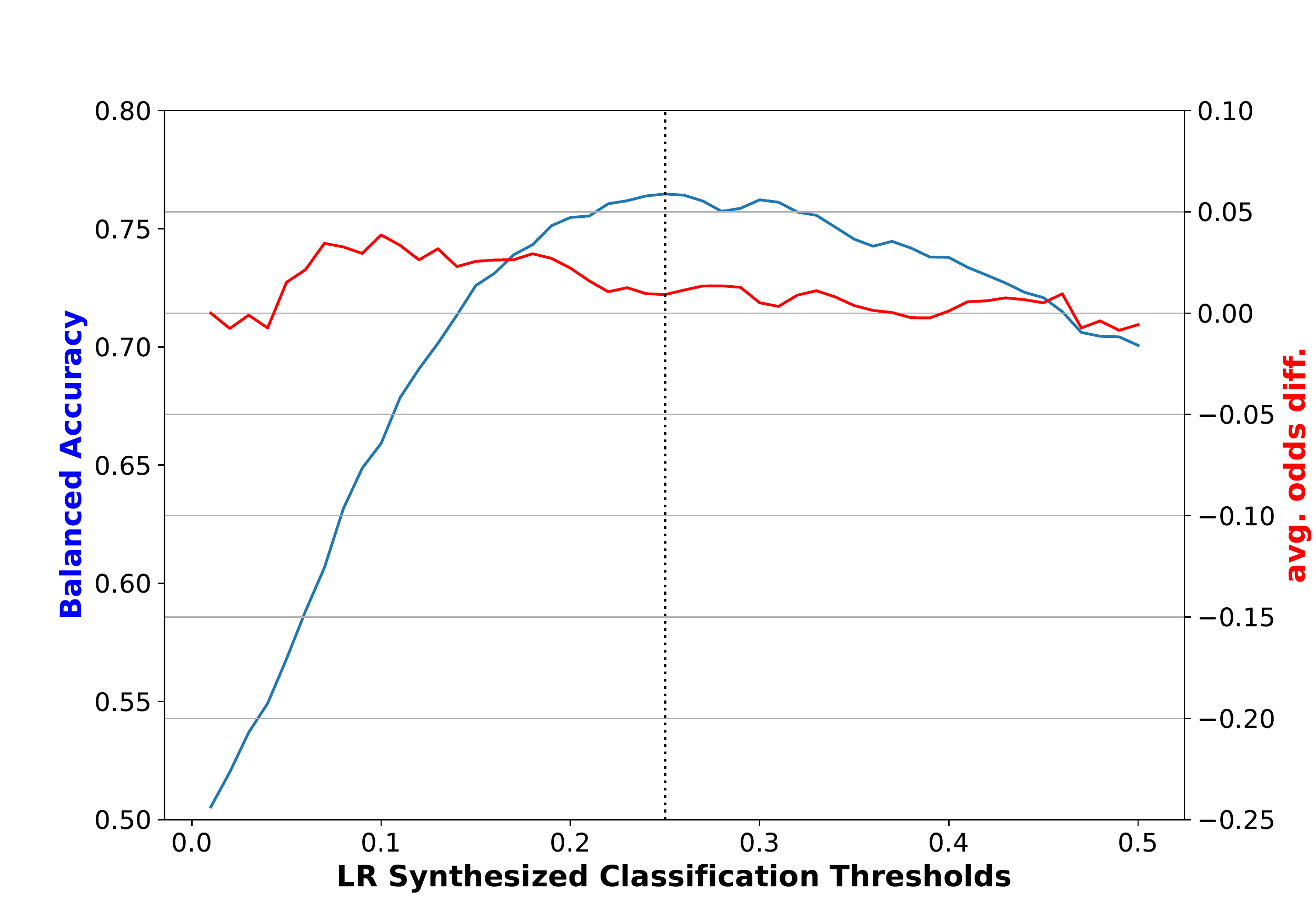}
\caption{\label{fig:LR_synth_odds} Accuracy and Avg. Odds Diff. for LR with Synthetic Data.}
\end{subfigure}
\newline
%\end{figure}
%\begin{figure}[!htb]
%\centering
\begin{subfigure}[b]{.225\textwidth}
\centering
\includegraphics[width=\textwidth]{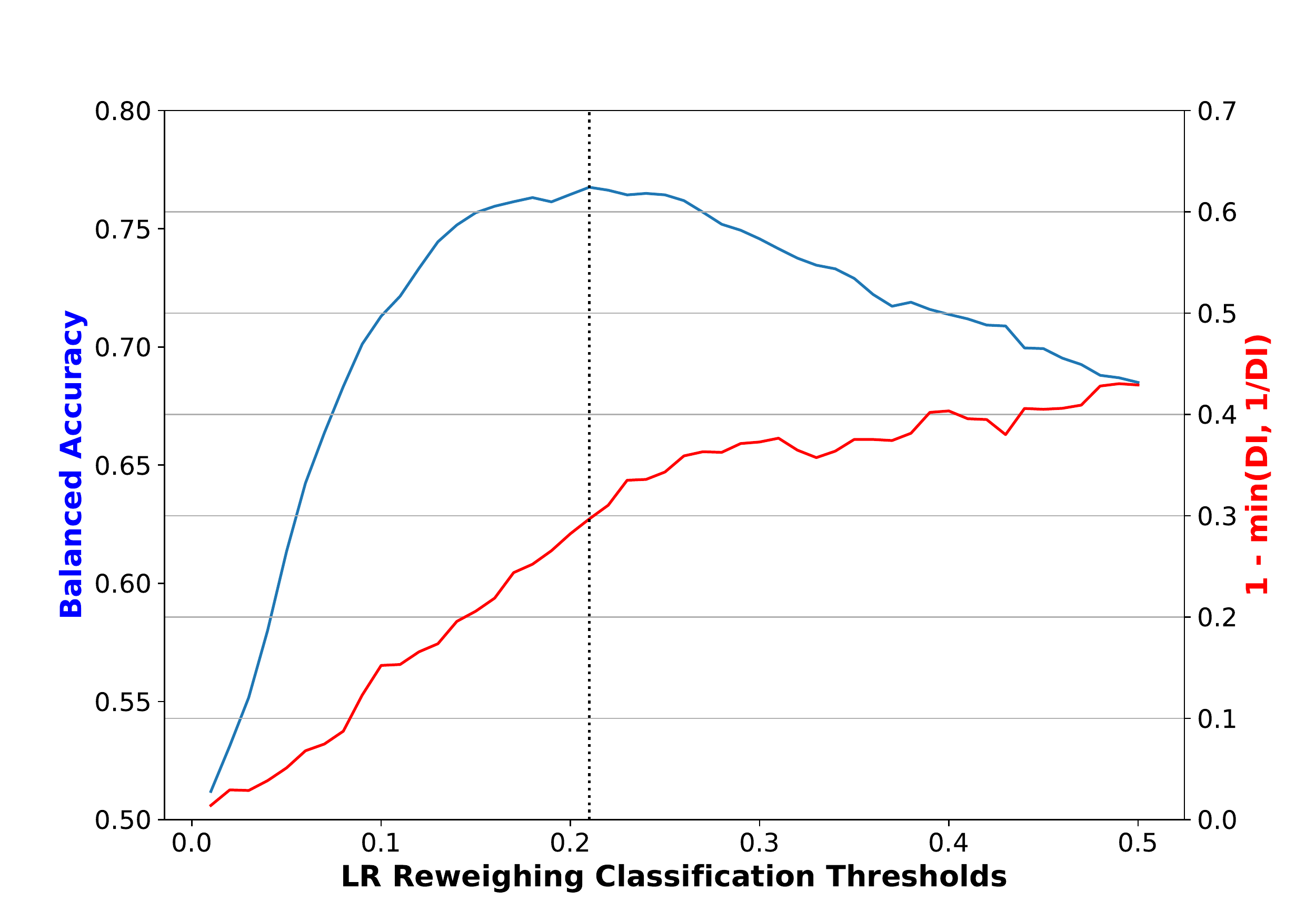}
\caption{\label{fig:LR_reweigh} Accuracy and Disparate Impact for LR with Reweighing.}
\end{subfigure}
\hfill
\begin{subfigure}[b]{.225\textwidth}
\centering
\includegraphics[width=\textwidth]{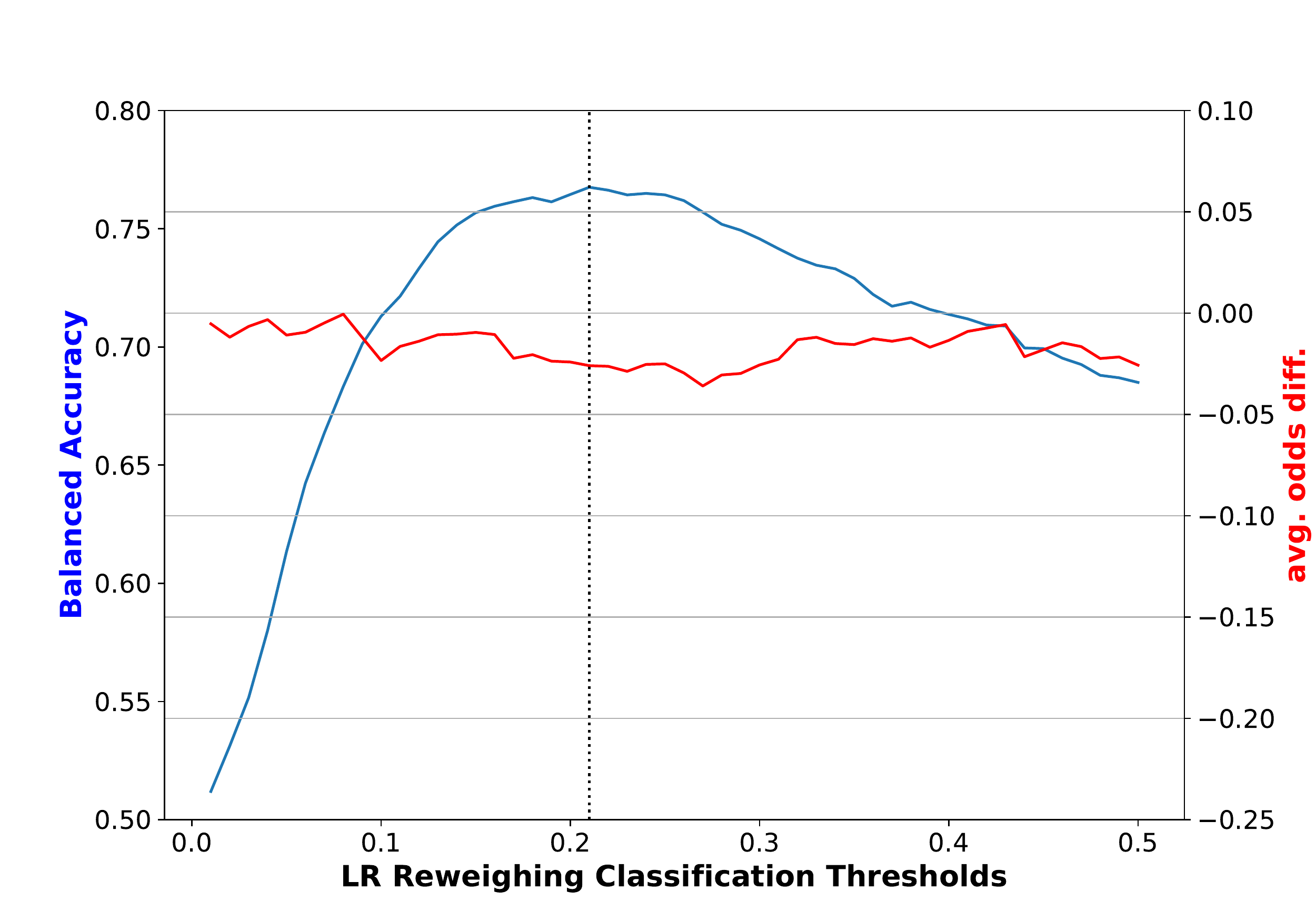}
\caption{\label{fig:LR_reweigh_odds} Accuracy \& Avg. Odds Diff. for LR with Reweighing.}
\end{subfigure}
\caption{\label{fig:adult_LR} Disparate Impact measure and Average Odds Difference of de-biasing techniques on Adult Data with LR.}
\end{figure}

\begin{figure}[!htb]
\centering
\begin{minipage}{.225\textwidth}
\centering
\includegraphics[width=\textwidth]{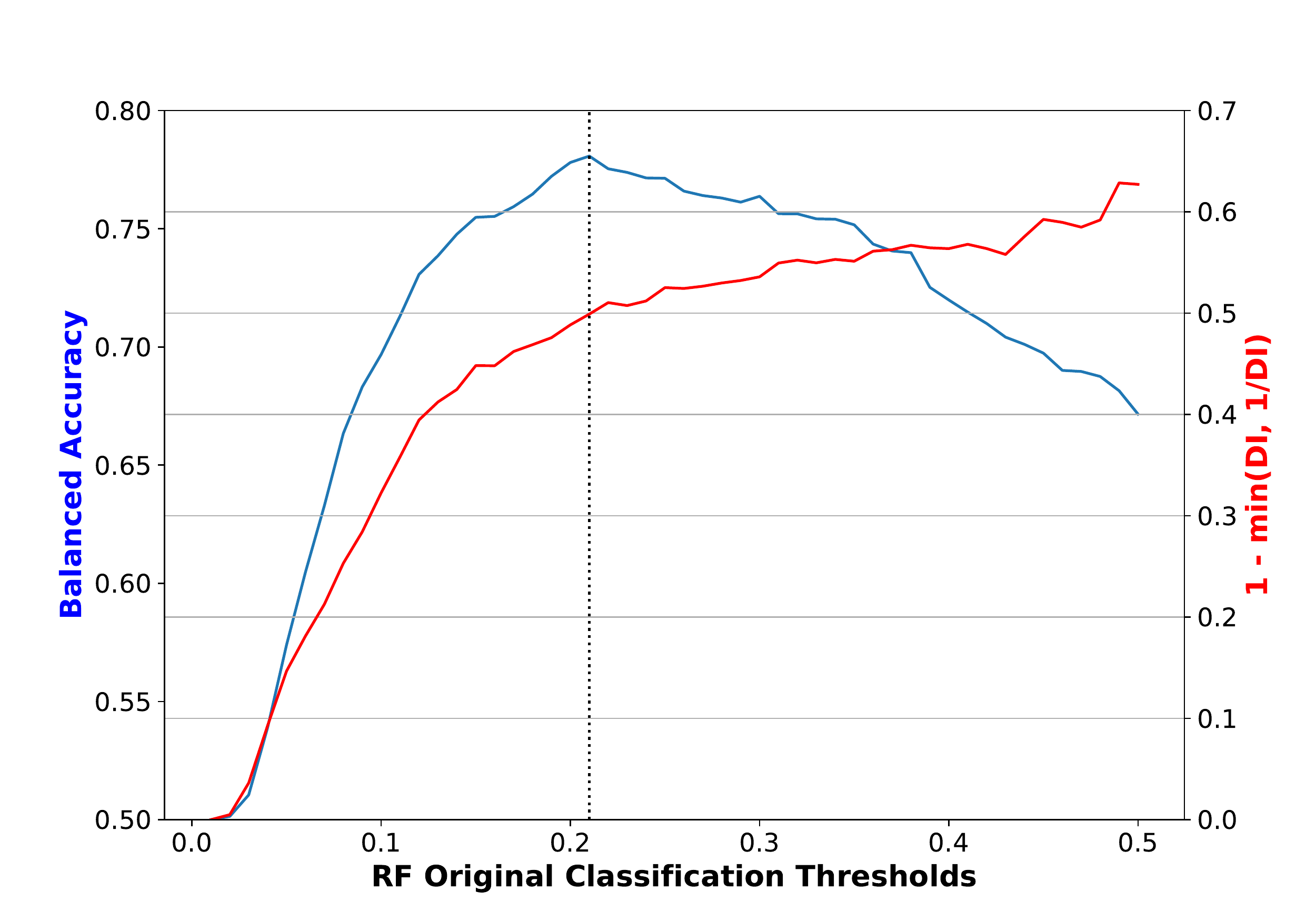}
%{\verb|RF_orig.eps|}
\end{minipage}
\begin{minipage}{.225\textwidth}
\centering
\includegraphics[width=\textwidth]{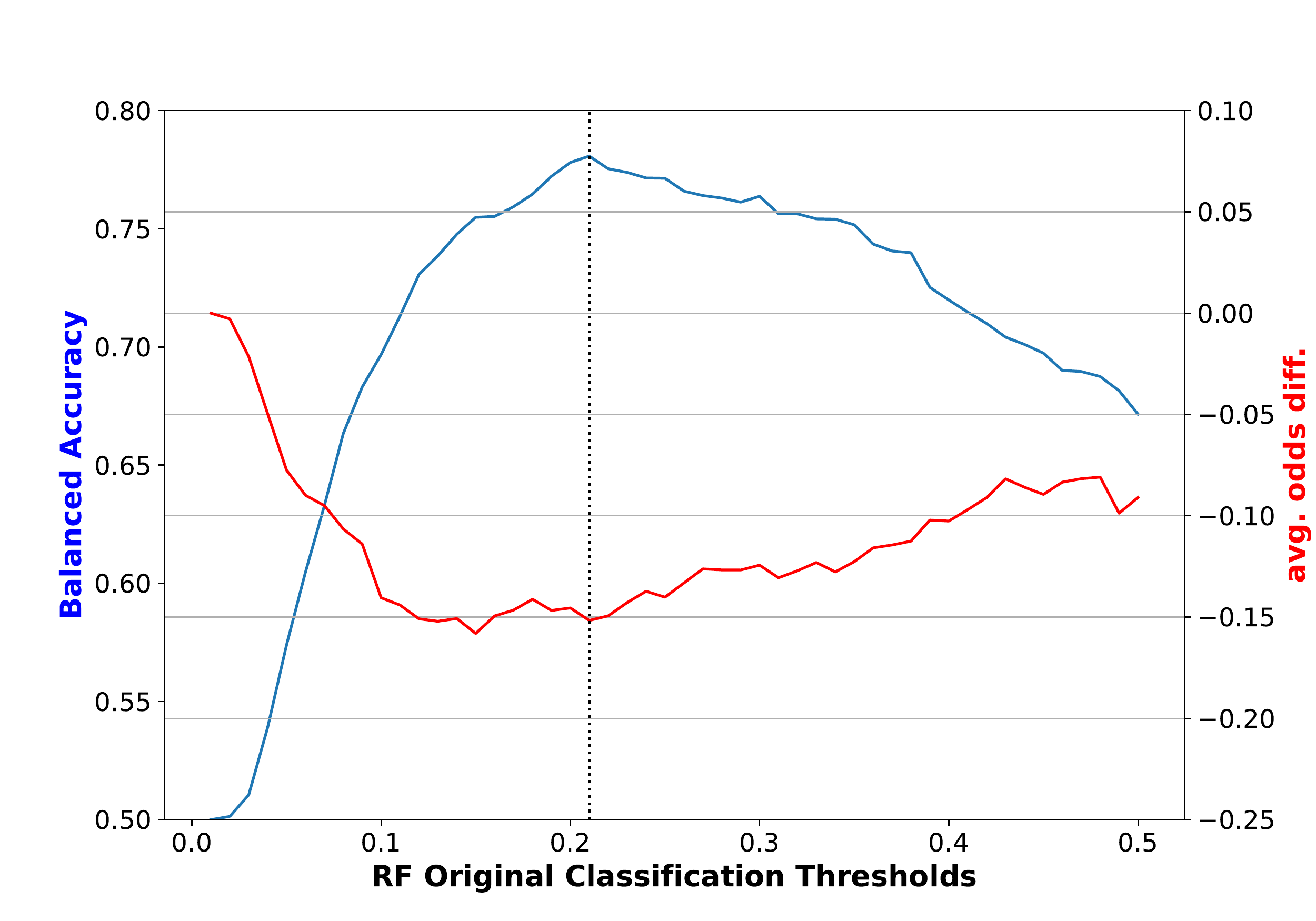}
%{\verb|RF_orig_odds.eps|}
\end{minipage}
%\end{figure}
%\begin{figure}[!htb]
%\centering
\begin{minipage}{.225\textwidth}
\centering
\includegraphics[width=\textwidth]{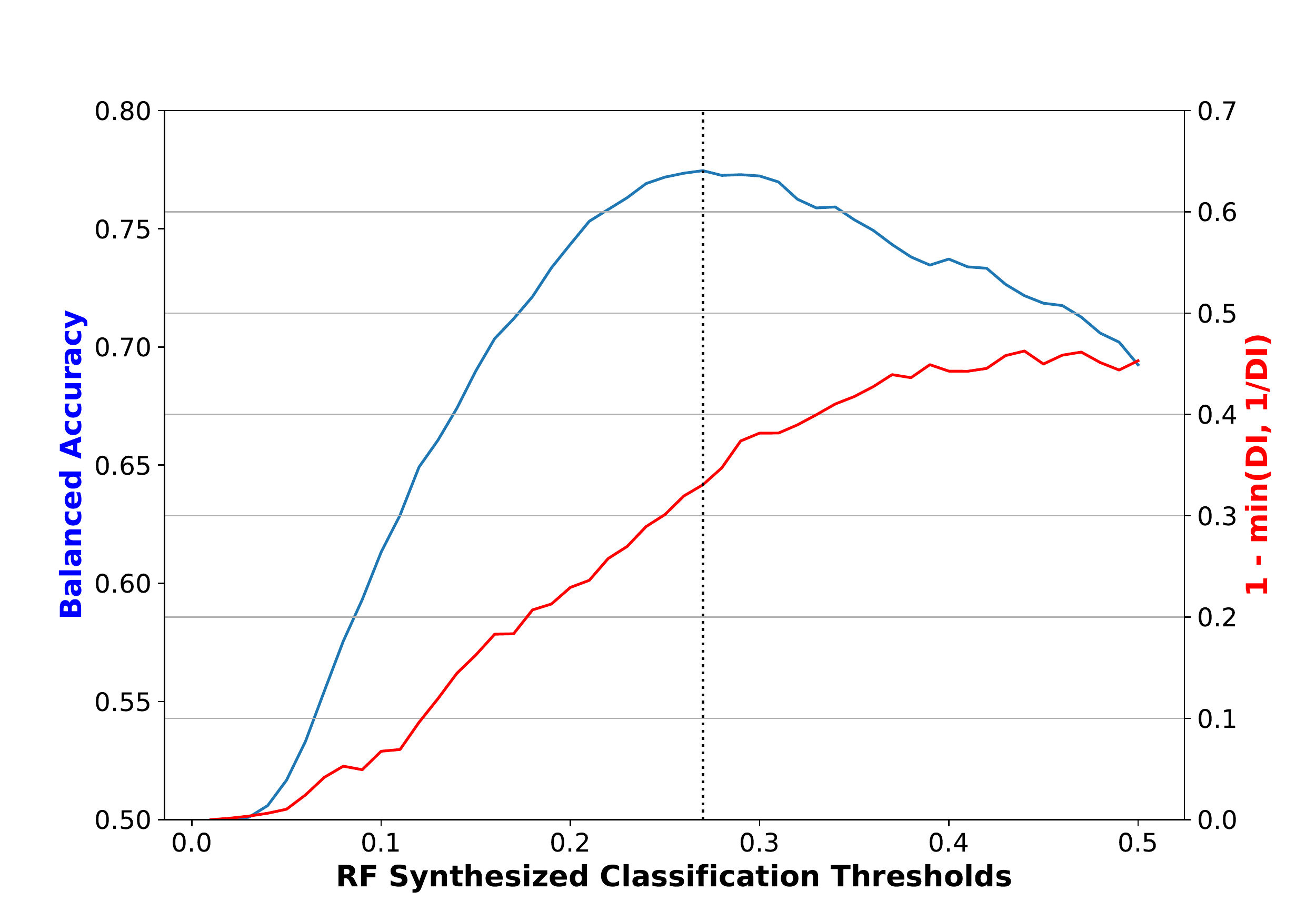}
%{\verb|RF_synth.eps|}
\end{minipage}%%
\begin{minipage}{.225\textwidth}
\centering
\includegraphics[width=\textwidth]{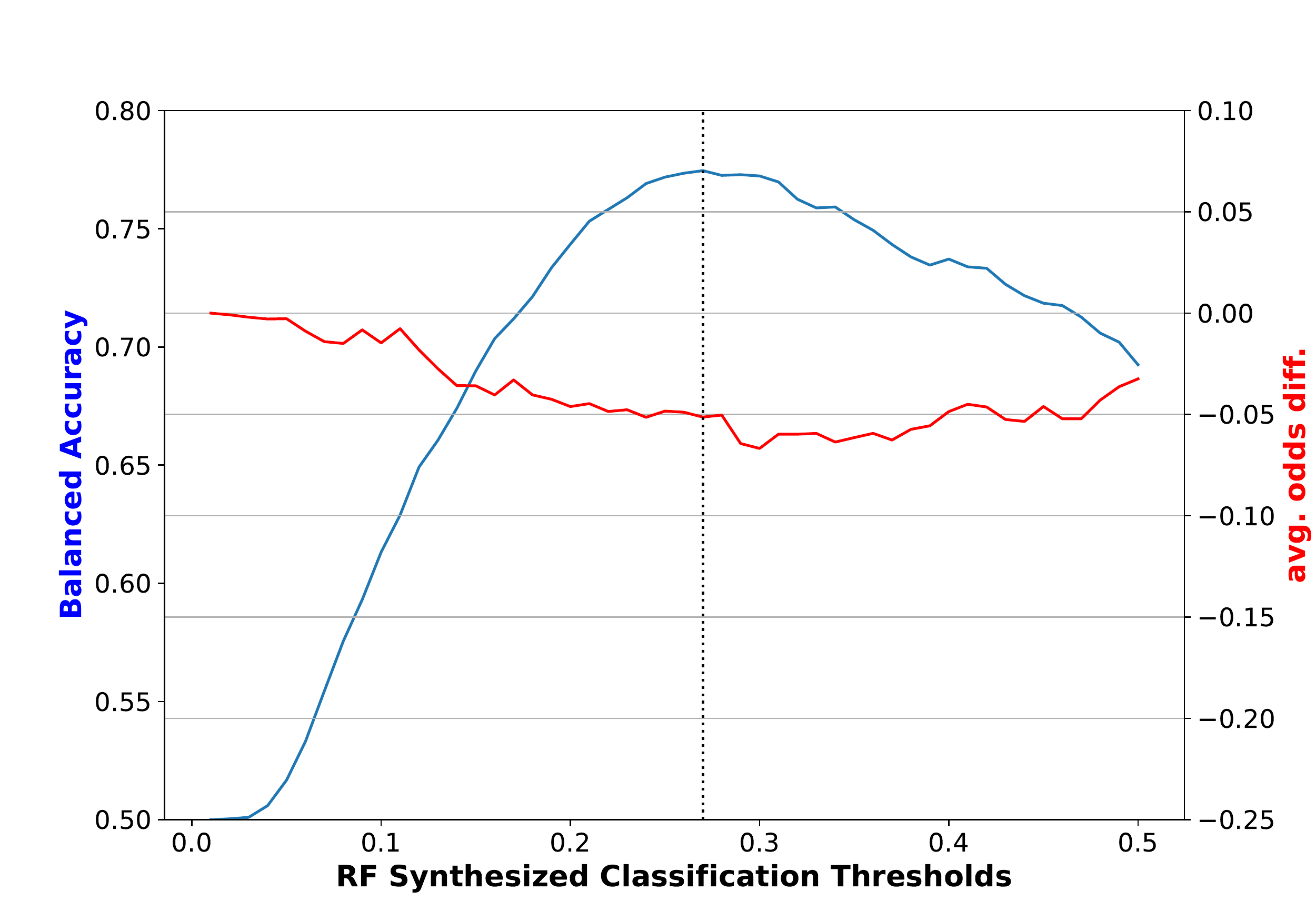}
%{\verb|RF_synth_odds.eps|}
\end{minipage}
%\end{figure}
%\begin{figure}[!htb]
%\centering
\begin{minipage}{.225\textwidth}
\centering
\includegraphics[width=\textwidth]{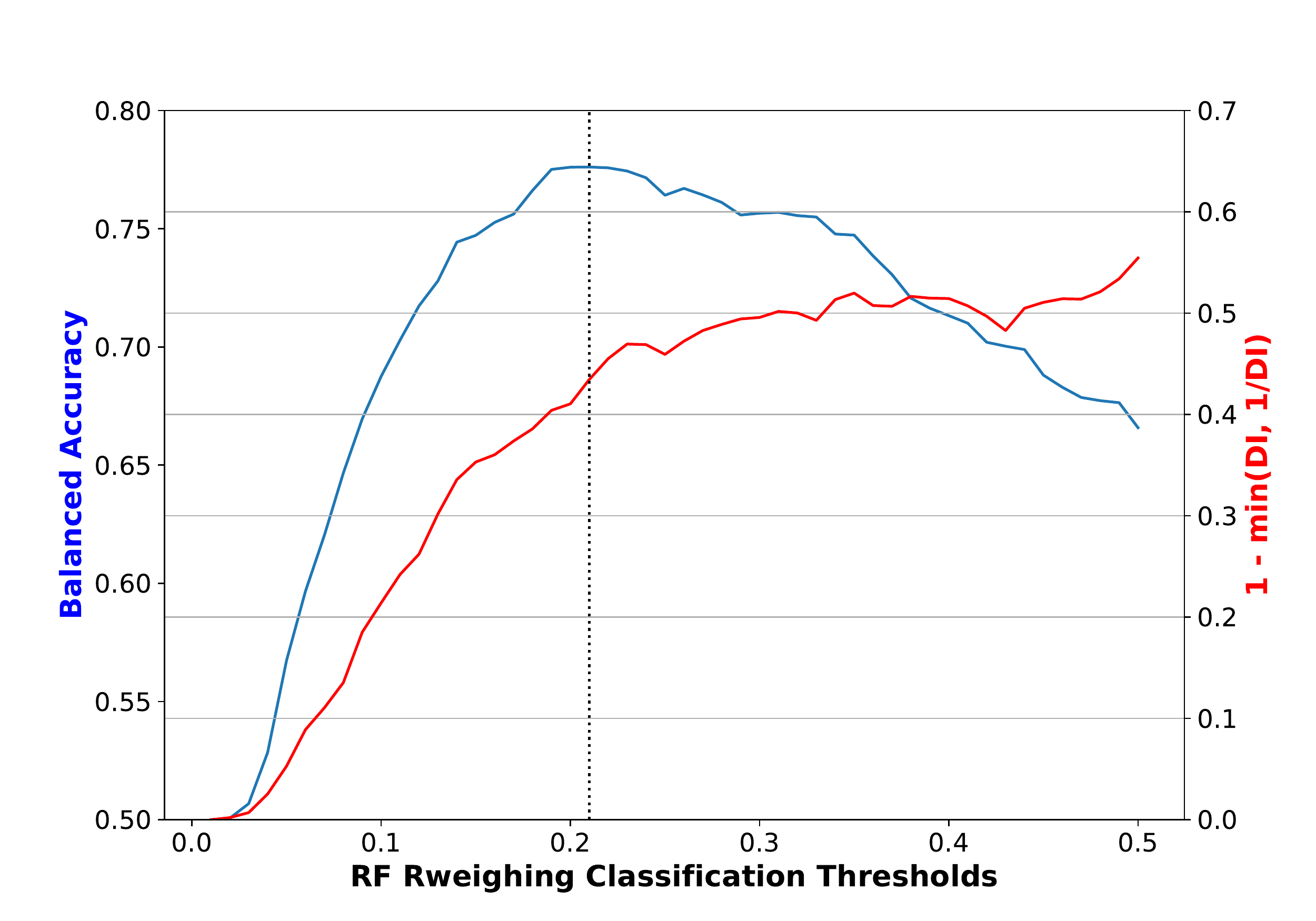}
%{\verb|RF_reweigh.eps|}
\end{minipage}%%
\begin{minipage}{.225\textwidth}
\centering
\includegraphics[width=\textwidth]{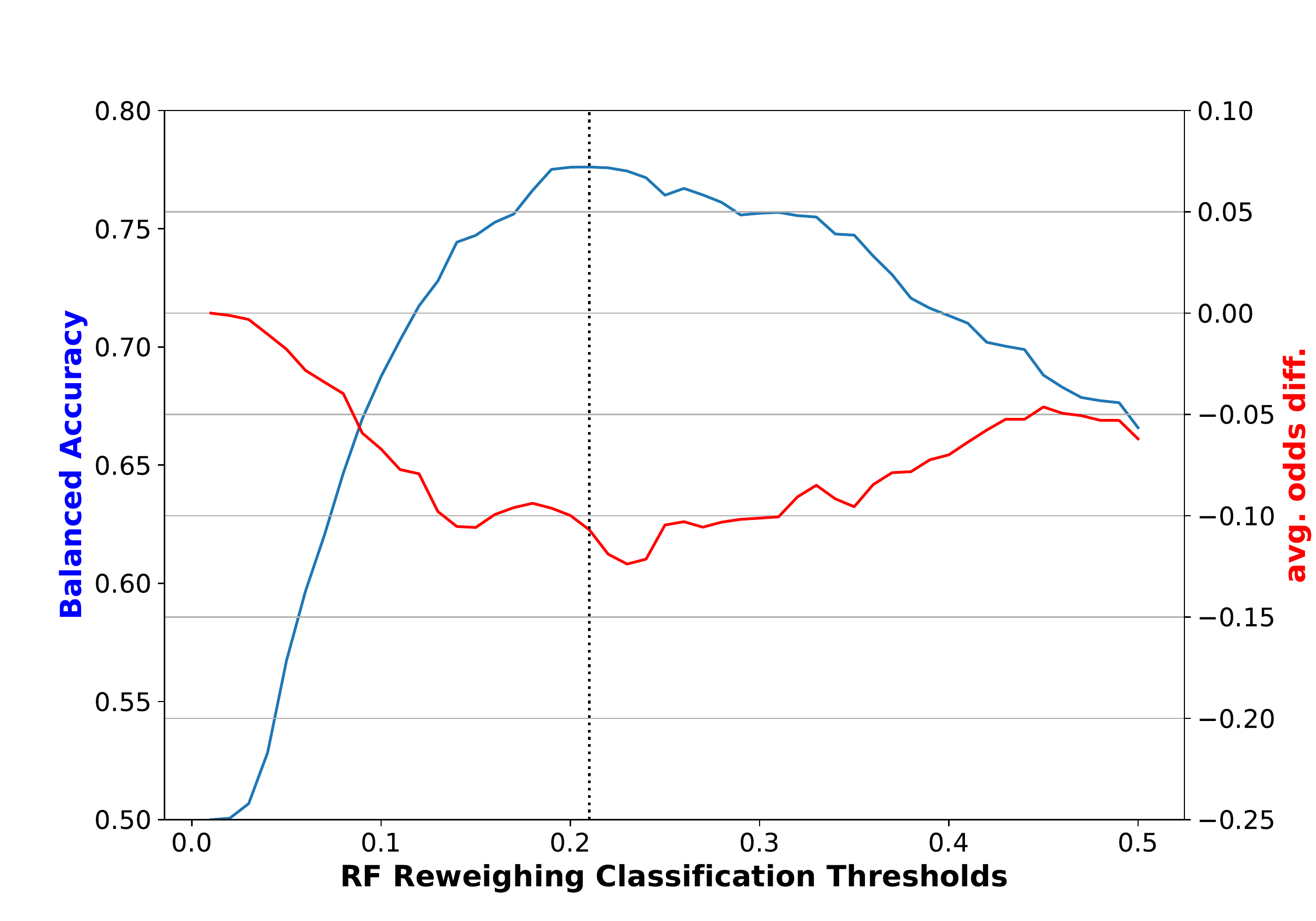}
%{\verb|RF_reweigh_odds.eps|}
\end{minipage}
\caption{\label{fig:adult_RF} Disparate Impact measure and Average Odds Difference of de-biasing techniques on Adult Data with RF.}
\end{figure}

%MK: I think in each dataset, we need to specify what is unfavored, what is favored, what is privileged etc.
%Yan: added
In the {\em Adult} dataset, the privileged group is {\em Male} and the favored class is {\em `$>$50K'}. 
Figure~\ref{fig:adult_LR} shows the accuracy and the fairness metrics: {\em disparate impact (DI) and average odds difference (AOD)} on the {\em Adult} data as the classification threshold increases from 0 to 0.5. Figure~\ref{fig:LR_orig}  shows the accuracy (blue) and the disparate impact measure $di = 1-\min(DI, \frac{1}{DI})$ (red), and Figure~\ref{fig:LR_orig_odds} shows the accuracy (blue) and AOD (red). Figures~\ref{fig:LR_synth} and~\ref{fig:LR_synth_odds} show the results of our de-biasing technique and Figures~\ref{fig:LR_reweigh} and~\ref{fig:LR_reweigh_odds} show the results of the {\em Reweighing} de-biasing technique. As can be observed, our de-biasing technique has significantly improved the disparate impact measure (0.2) compared to the original LR model (0.6), and outperforms the {\em Reweighing} technique (0.3) at the balanced accuracy (vertical line). Similarly, we observed better fairness measure (0.01) in terms of the absolute value of the average odds difference  than the original LR model (0.2) and the {\em Reweighing} technique (0.025). When the baseline algorithm is {\em Random Forest} (RF), we observed similar results in Figure~\ref{fig:adult_RF}). 

%%%%%%%%%%%%%%%%%%%
%\begin{comment}
%\end{comment}
%%%%%%%%%%%%%%%%%%%%%%%%%%

\begin{figure}[!htb]
\centering
\begin{minipage}{.225\textwidth}
\centering
\includegraphics[width=\textwidth]{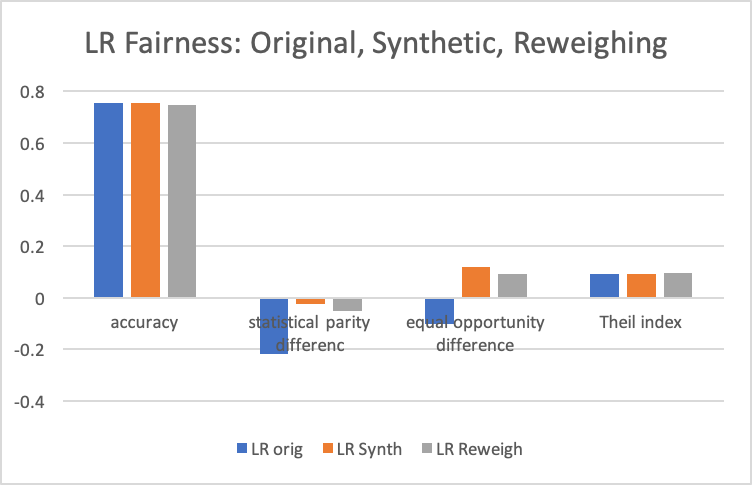}
%{\verb|LR_Stat|}
\end{minipage}
\hfill
\begin{minipage}{.24\textwidth}
\centering
\includegraphics[width=\textwidth]{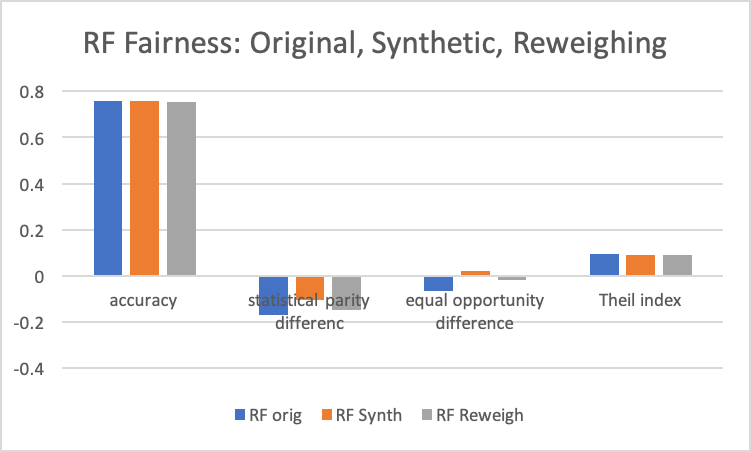}
%{\verb|RF_Stat|}
\end{minipage}
%\begin{minipage}{.225\textwidth}
%\centering
%\includegraphics[trim=1cm 20cm 1cm 1cm, clip, width=\textwidth]{stats.pdf}
%{\verb|Stat_medical|}
%\end{minipage}%%
%%\end{figure}
%%\begin{figure}[!htb]
%%\centering
%\begin{minipage}{.225\textwidth}
%\centering
%\includegraphics[trim=1cm 20cm 1cm 1cm, clip, width=\textwidth]{stats_adult.pdf}
%{\verb|Stat_Adult|}
%\end{minipage}
\caption{\label{fig:adult_morestat} Results of additional fairness measure on Adult data.}
\end{figure}

%MK: please comment on what the figure~\ref{fig:adult_morestat} means. do we see similar results??
The results on other fairness measures: {\em statistical parity difference}, {\em equal opportunity difference}, and {\em Theil index}, are shown in Figure~\ref{fig:adult_morestat} when the baseline learning algorithm is LR and RF,  respectively. Our technique improves fairness and is  better than {\em Reweighing} in general. 

\subsection{Compas Data}
\label{subsec:compas}
\begin{figure}[!htb]
\centering
\includegraphics[trim=0cm 0cm 0cm 0cm, clip, width=0.45\textwidth]{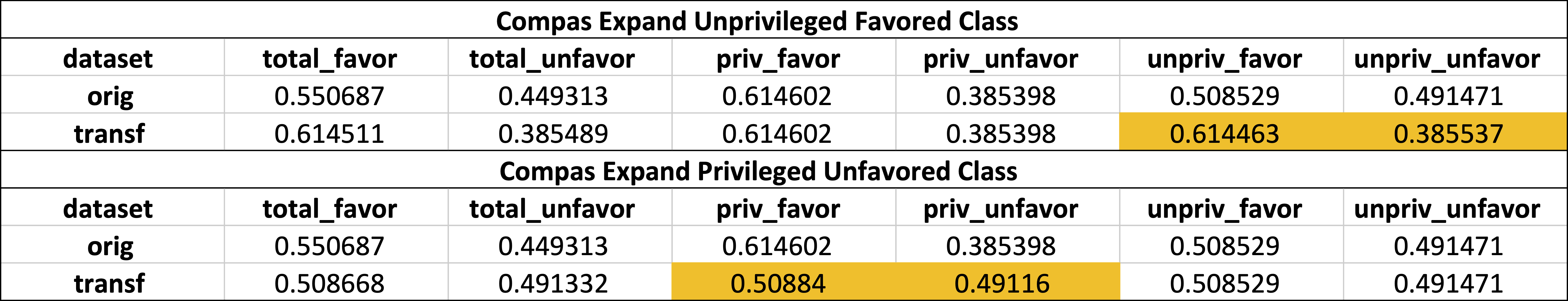}
\caption{\label{fig:compas_prior}The prior distributions of the favored and the unfavored classes before and after synthetic data augmentation.}
%{\verb|Compas Favored/Unfavored Distribution|}
\end{figure}

%MK: Next sentence is not clear. Higher in what sense??
%MK: I think in each dataset, we need to specify what is unfavored, what is favored, what is privileged etc.
%Yan: Done
In the {\em Compas} dataset, the number of examples in the favored class ({\em no recidivism}) is approximately 10\% higher than in the unfavored class, and the gap is wider (approx. 22\%) between the privileged group ({\em Caucasian}) and the unprivileged group, as shown in Figure~\ref{fig:compas_prior} (rows labeled {\bf orig}). We generate synthetic favored samples in the underrepresented group ({\em Not Caucasian}) to reduce the difference in the positive base rate between the privileged and the unprivileged groups. The distribution of favored and unfavored classes for the underrepresented group after oversampling is highlighted and shown in Figure~\ref{fig:compas_prior} (row labeled {\bf transf} under {\bf Compas Expand Unprivileged Favored Class}). We also oversample unfavored class for the privileged group (row labeled {\bf transf} under {\bf Compas Expand Privileged Unfavored Class}).

\begin{figure}[!htb]
\centering
%\begin{minipage}{.595\textwidth}
%\centering
%\includegraphics[width=\textwidth]{majority_priv.png}
%{\verb|German Credit|}
%\end{minipage}
%\par\bigskip
%\begin{minipage}{\textwidth}
%\centering
\includegraphics[trim=3cm 0.5cm 2cm 2cm, clip, width=0.35\textwidth]{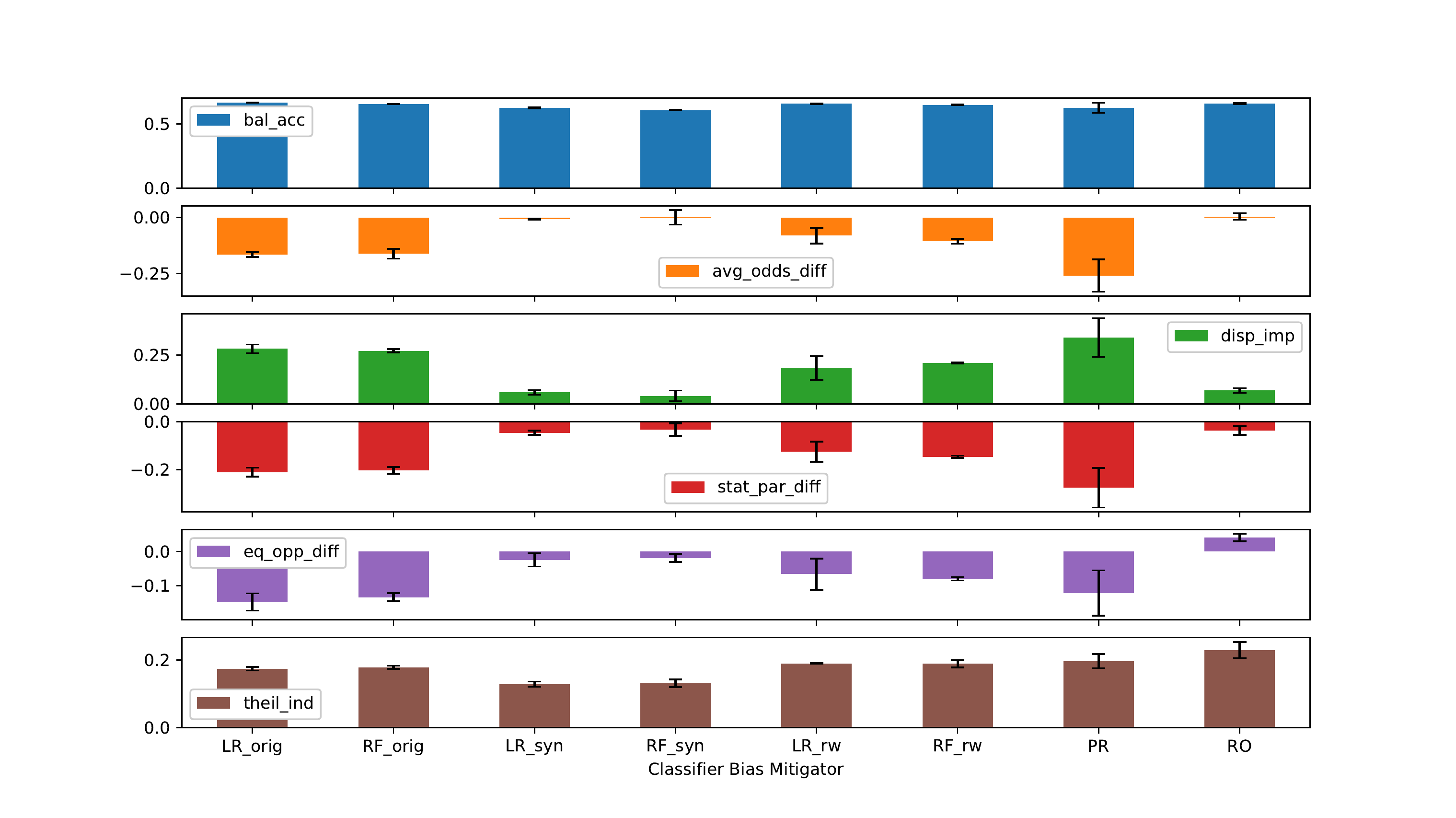}
\caption{\label{fig:compas_upfv} Fairness results on Compas: oversample unprivileged favored class.}
%\end{minipage}
\end{figure}

\begin{figure}[!htb]
\centering
%\begin{minipage}{.595\textwidth}
%\centering
%\includegraphics[width=\textwidth]{majority_priv.png}
%{\verb|German Credit|}
%\end{minipage}
%\par\bigskip
%\begin{minipage}{\textwidth}
%\centering
\includegraphics[trim=3cm 0.5cm 2cm 1.5cm, clip, width=0.35\textwidth]{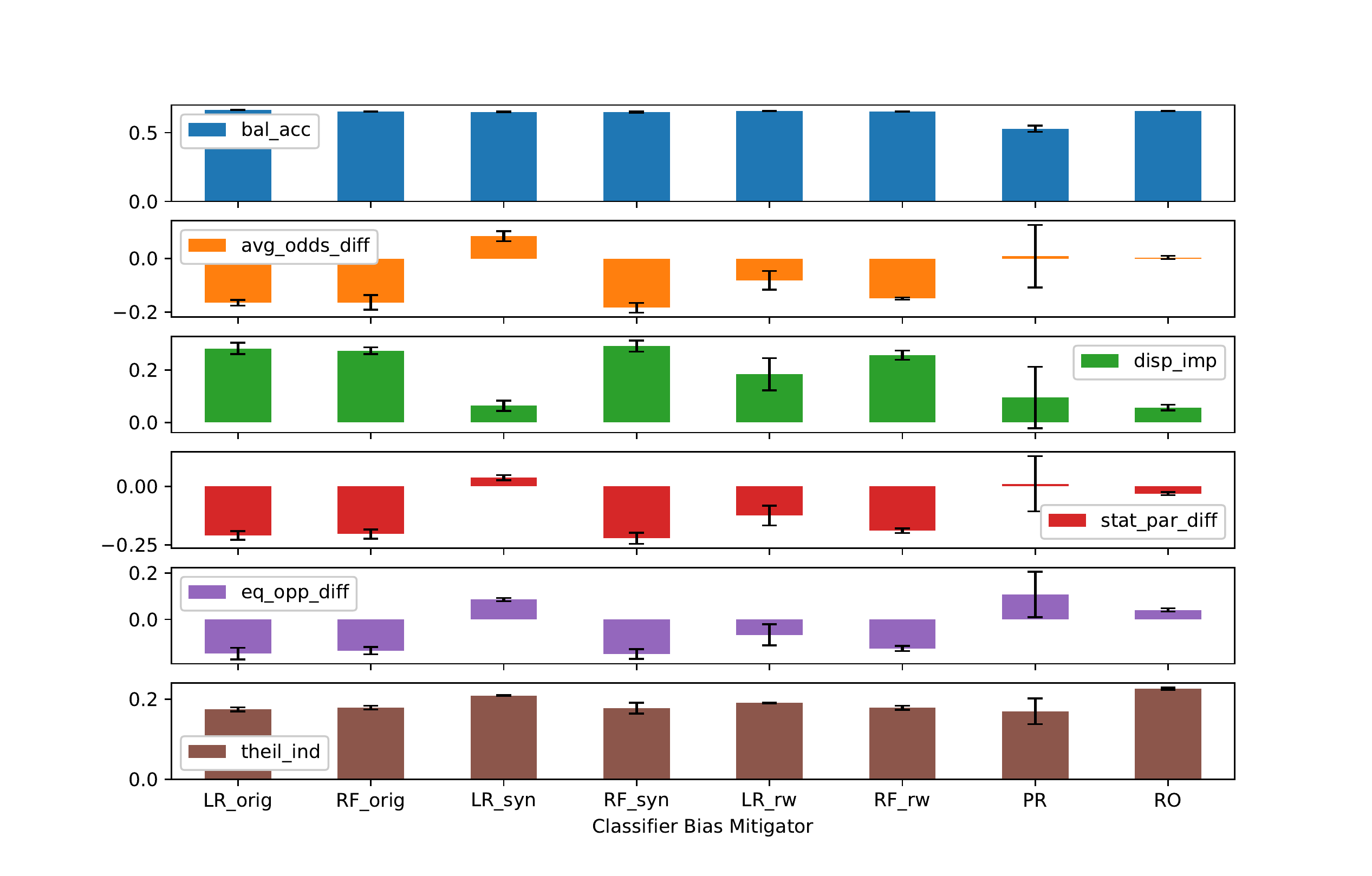}
%{\verb|Compas Expand Privileged Unfavored|}
\caption{\label{fig:compas_pufv} Fairness results on Compas: oversample privileged unfavored class.}
%\end{minipage}
\end{figure}

%Figures~\ref{fig:comp_dist} and~\ref{fig:mep_dist} show the distributions of favored and unfavoed class distributions in the privileged and unprivileged groups in the Compas and the Medical Expense datasets. 
Figures~\ref{fig:compas_upfv} and~\ref{fig:compas_pufv} illustrate, from the top to the bottom row, the {\em balanced accuracy},  {\em average odds difference}, {\em disparate impact}, {\em statistical parity difference}, {\em equal opportunity difference}, and {\em Theil index} for the two cases. Our results are compared to other de-biasing techniques and displayed column wise, from left to right, the baseline learning algorithm, our synthetic data de-biasing technique applied to LR and RF, {\em Reweighing} (pre-processing) applied to LR and RF, {\em Prejudice Remover} (in-processing) and  {\em Reject Option} (post-processing) both applied only to LR as implemented in the AI Fairness 360 library. Our  de-biasing technique demonstrates better overall fairness measure when we oversample the favored class for the underrepresented group. All the experiments were run 10 times and the averaged results (boxes) and the standard errors (error bars) were reported.
%%%%%%%%%%%%%%%%%%%%%%%%%%%%%%%%%%%%%%%%%%%
\begin{comment}
%%%%%%%%%%%%%%%%%%%%%%%%%%%%%%%%%%%%%%%%%%%
\begin{figure}[!htb]
\centering
\begin{minipage}{.45\textwidth}
\centering
%\includegraphics[trim=1cm 20cm 1cm 1cm, clip, width=\textwidth]{Compas.png}
\includegraphics[width=\textwidth]{Compas.png}
{\verb|Compas|}
\end{minipage}
%\end{figure}
%\begin{figure}[!htb]
%\centering
\begin{minipage}{.45\textwidth}
\centering
\includegraphics[width=\textwidth]{Compas_op.png}
{\verb|Compas_OP|}
\end{minipage}
\end{figure}
\begin{figure}[!htb]
\centering
\begin{minipage}{.45\textwidth}
\centering
\includegraphics[width=\textwidth]{German.png}
{\verb|German Credit|}
\end{minipage}
\begin{minipage}{.45\textwidth}
\centering
\includegraphics[width=\textwidth]{german_foreign_2.png}
{\verb|German Credit|}
\end{minipage}
\begin{minipage}{.45\textwidth}
\centering
\includegraphics[width=\textwidth]{german_foreign_6.png}
{\verb|German Credit|}
\end{minipage}
\begin{minipage}{.45\textwidth}
\centering
\includegraphics[width=\textwidth]{german_foreign_10.png}
{\verb|German Credit|}
\end{minipage}
\end{figure}
%%%%%%%%%%%%%%%%%%%%%%%%%%%%%%%%%%%%%%%%%%%
\end{comment}
%%%%%%%%%%%%%%%%%%%%%%%%%%%%%%%%%%%%%%%%%%%
%\clearpage

The results are less impressive when we oversample the unfavored class for the privileged group, as shown in Figure~\ref{fig:compas_pufv}. Our de-biasing  technique tend to work better when the class prior of the original dataset is respected. Also notice that {\em Prejudice Remover} tends to produce large variances, and there is a clear trade-off between fairness and accuracy. 

\begin{figure}[!htb]
\centering
\includegraphics[trim=2.5cm 0.5cm 2cm 1.5cm, clip,width=0.35\textwidth]{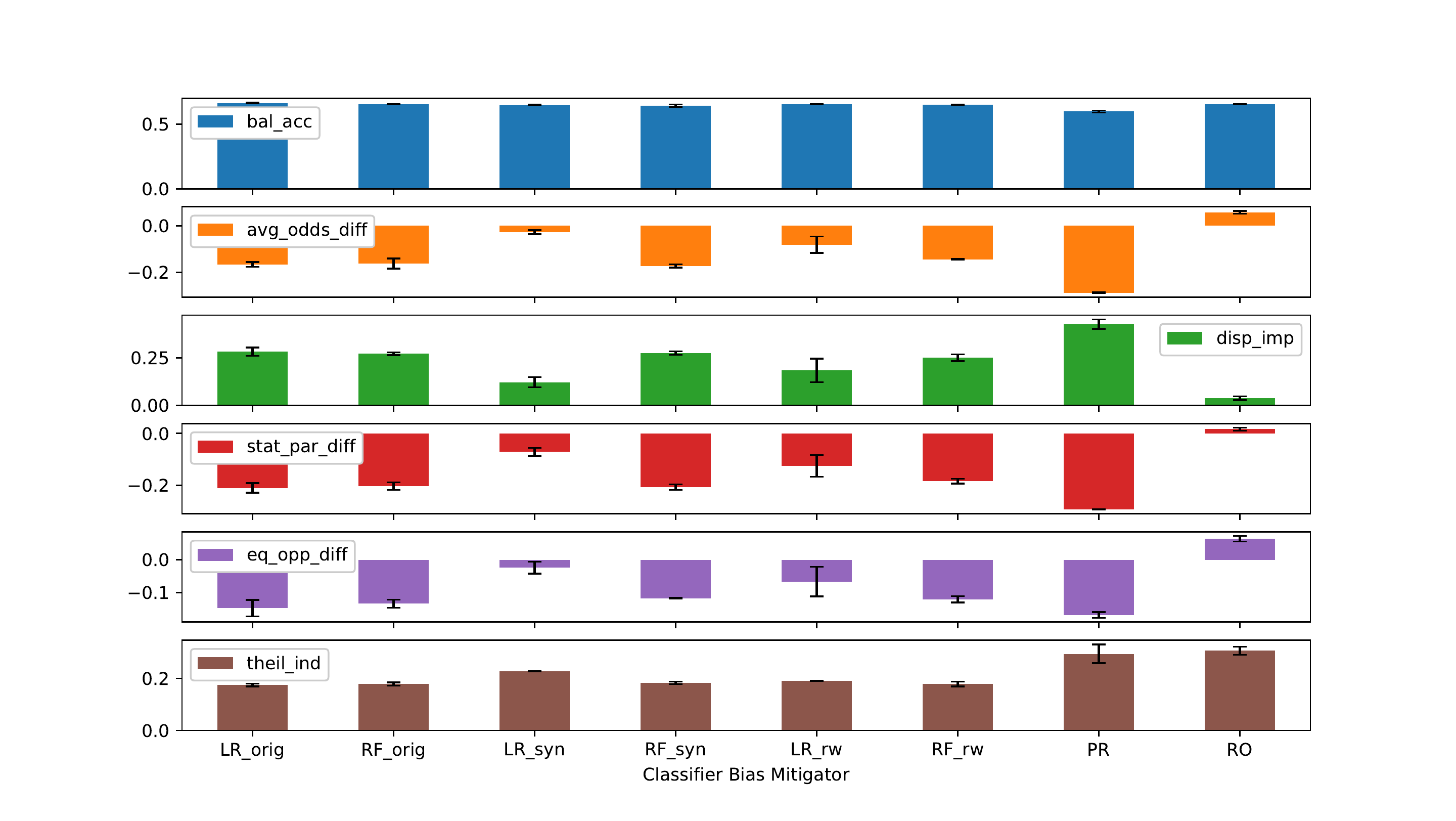}
%{\verb|Compas Expand Privileged Balanced|}
\caption{\label{fig:compas_balance} Fairness results on Compas: oversample both unprivileged favored class and privileged unfavored class.}
\end{figure}

We also tried to combine the two cases, that is, oversample both the favored class for the underrepresented group and the unfavored class for the privileged group. The results of the combined case lie in between the first and the second cases, as shown in Figure~\ref{fig:compas_balance}. 

\subsection{German Credit Data}
\label{subsec:german}

\begin{figure}[!htb]
\centering
\includegraphics[width=0.45\textwidth]{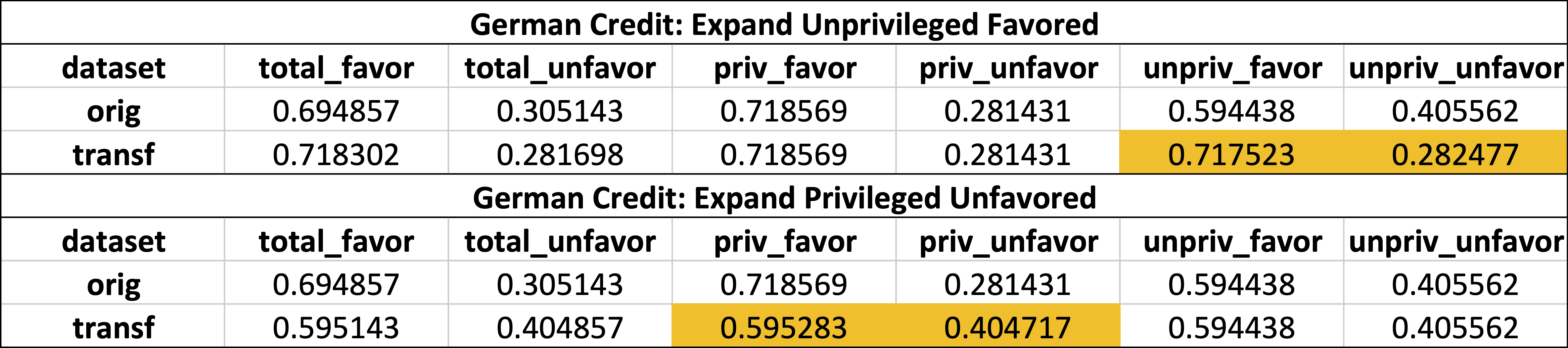}
\caption{\label{fig:german_prior}German-credit Favored/Unfavored Distribution.}
\end{figure}

The distribution of the favored ({\em Good Credit}) and unfavored ({\em Bad Credit}) classes of the {\em German Credit} data before and after oversampling is shown in Figure~\ref{fig:german_prior}. The privileged group is {\em Old} and the unprivileged group is {\em Young}.

\begin{figure}[!htb]
\centering
\begin{subfigure}{.305\textwidth}
\includegraphics[trim=2.5cm 0.5cm 2cm 1.5cm, clip,width=\textwidth]{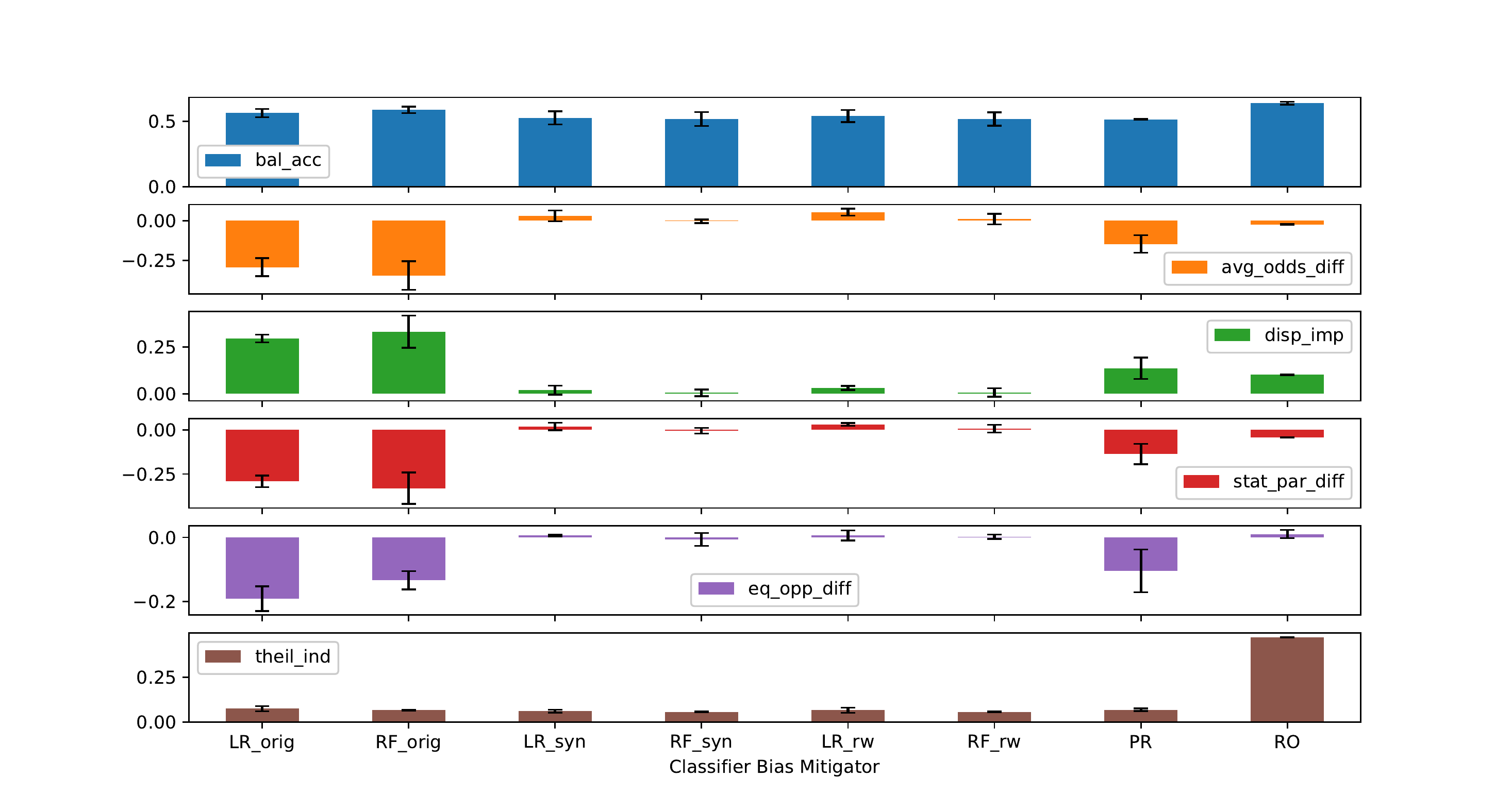}
\caption{\label{fig:german_case1} German Credit: Expand Unprivileged Favored.}
\end{subfigure}
\begin{subfigure}{.305\textwidth}
\centering
\includegraphics[trim=2.5cm 0.5cm 2cm 1.5cm, clip,width=\textwidth]{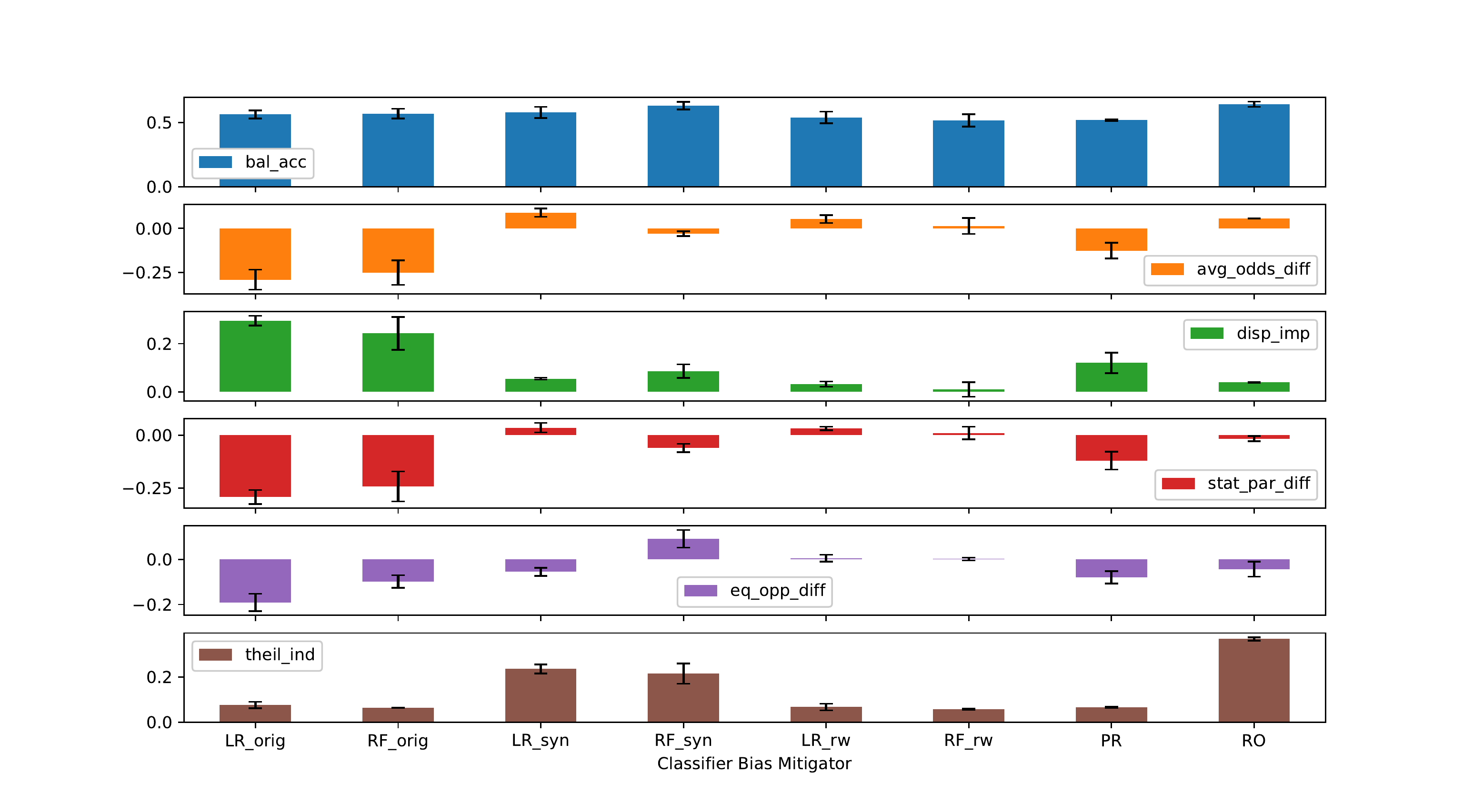}
\caption{\label{fig:german_case2} German Credit: Expand Privileged Unfavored.}
\end{subfigure}
\begin{subfigure}{.305\textwidth}
\centering
\includegraphics[trim=1cm 0cm 1cm 1cm, clip,width=\textwidth]{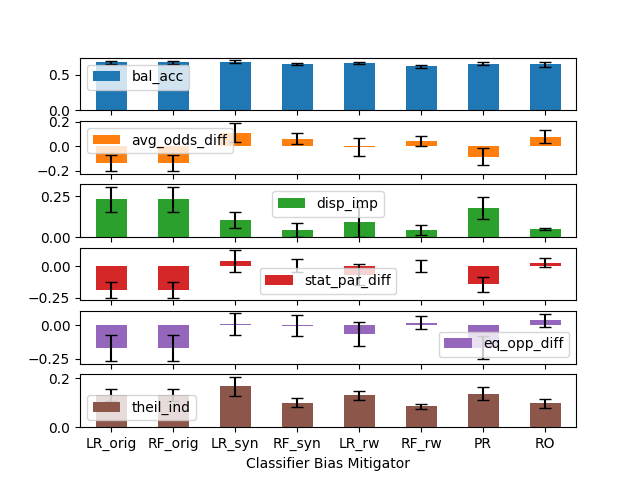}
\caption{\label{fig:german_case3} German Credit: Expand Unprivileged Favored and Privileged Unfavored.}
\end{subfigure}
\caption{\label{fig:german} Fairness results on German Credit data.}
\end{figure}

Our de-biasing technique is comparable to {\em Reweighing}, and overall outperforms {\em Prejudice Remover} (PR) and {\em Reject Option} (RO) when we oversample the favored class for the underrepresented group, but less effective when we oversample unfavored class for the privileged group, as shown in Figures~\ref{fig:german_case1} and~\ref{fig:german_case2}. The results of the combined case for our de-biasing technique again lie in between the two cases, as shown in Figure~\ref{fig:german_case3}. 

\subsection{Medical Expense Price Data}
\label{subsec:mep}

\begin{figure}[!htb]
\centering
\includegraphics[width=0.45\textwidth]{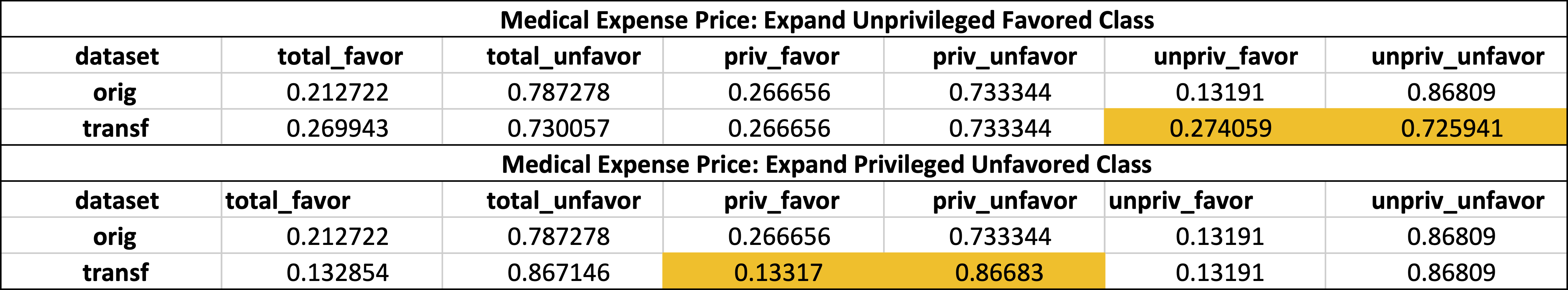}
\caption{\label{fig:mep_prior} Medical Expense Price Favored/Unfavored Distribution.}
\end{figure}

The favored/unfavored class distribution of the {\em Medical Expense Price} data before and after oversampling is shown in Figure~\ref{fig:mep_prior}. The favored class is {\em `$\ge$ 10 Visits'} and the privileged group is the {\em White} race. Notice unlike in other datasets, the majority class is the unfavored class in this dataset. 
%\clearpage

Our de-biasing technique is comparable to {\em Reweighing}, {\em Prejudice Remover} (PR), and {\em Reject Option} (RO) in both cases when we oversample the favored class for the underrepresented group (case 1) and the unfavored class for the privileged group (case 2). Again, {\em Prejudice Remover} shows a clear trade-off between fairness and accuracy, and  its {\em Theil index} is significantly worse than other de-biasing techniques. 
\begin{figure}[!htb]
\centering
\begin{subfigure}{.305\textwidth}
\includegraphics[trim=2.5cm 0.5cm 2cm 1.5cm, clip,width=\textwidth]{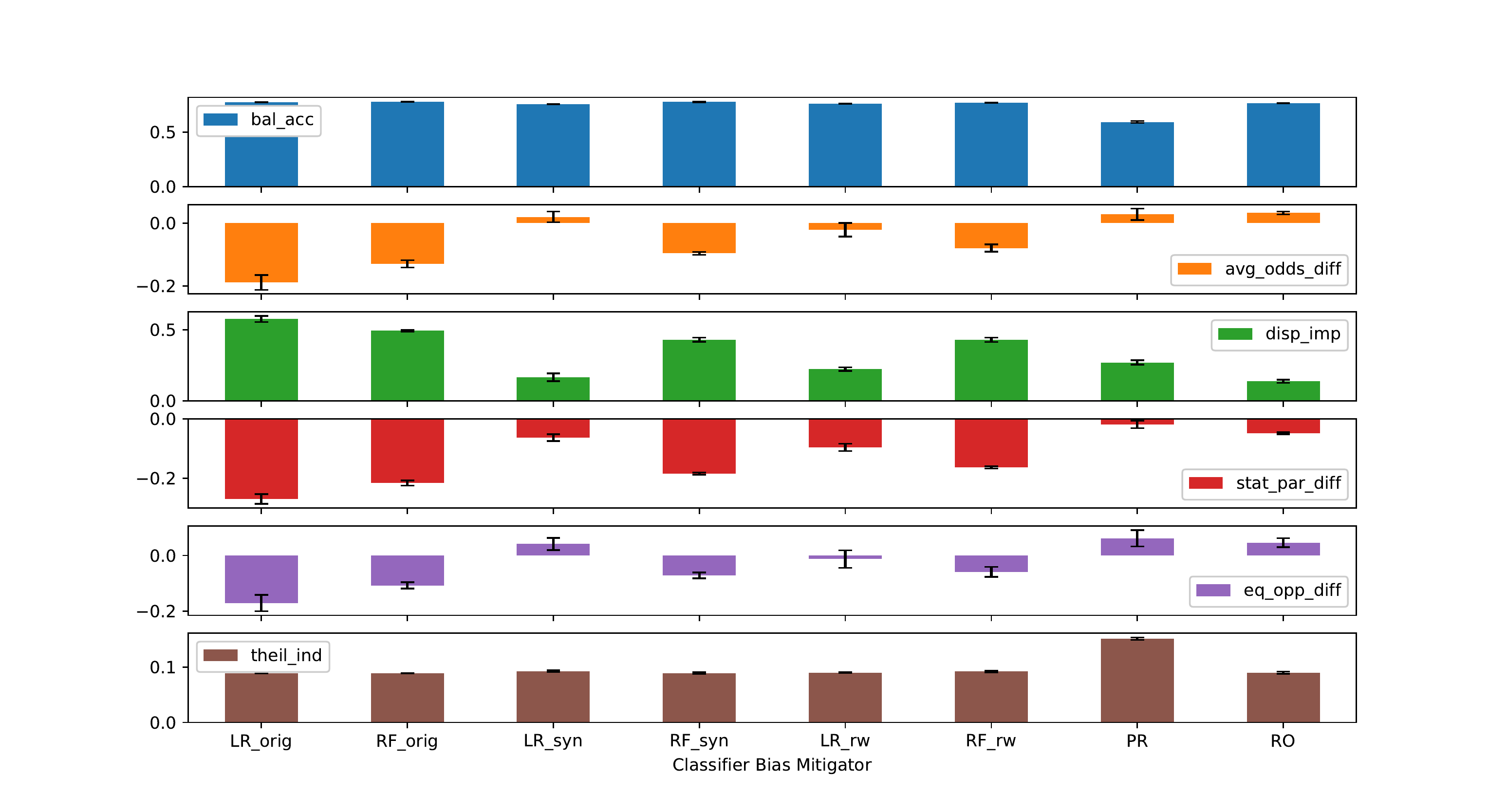}
\caption{\label{fig:mep_case1} Medical Expense Price: Expand Unprivileged Favored.}
\end{subfigure}
\begin{subfigure}{.305\textwidth}
\centering
\includegraphics[trim=2.5cm 0.5cm 2cm 1.5cm, clip,width=\textwidth]{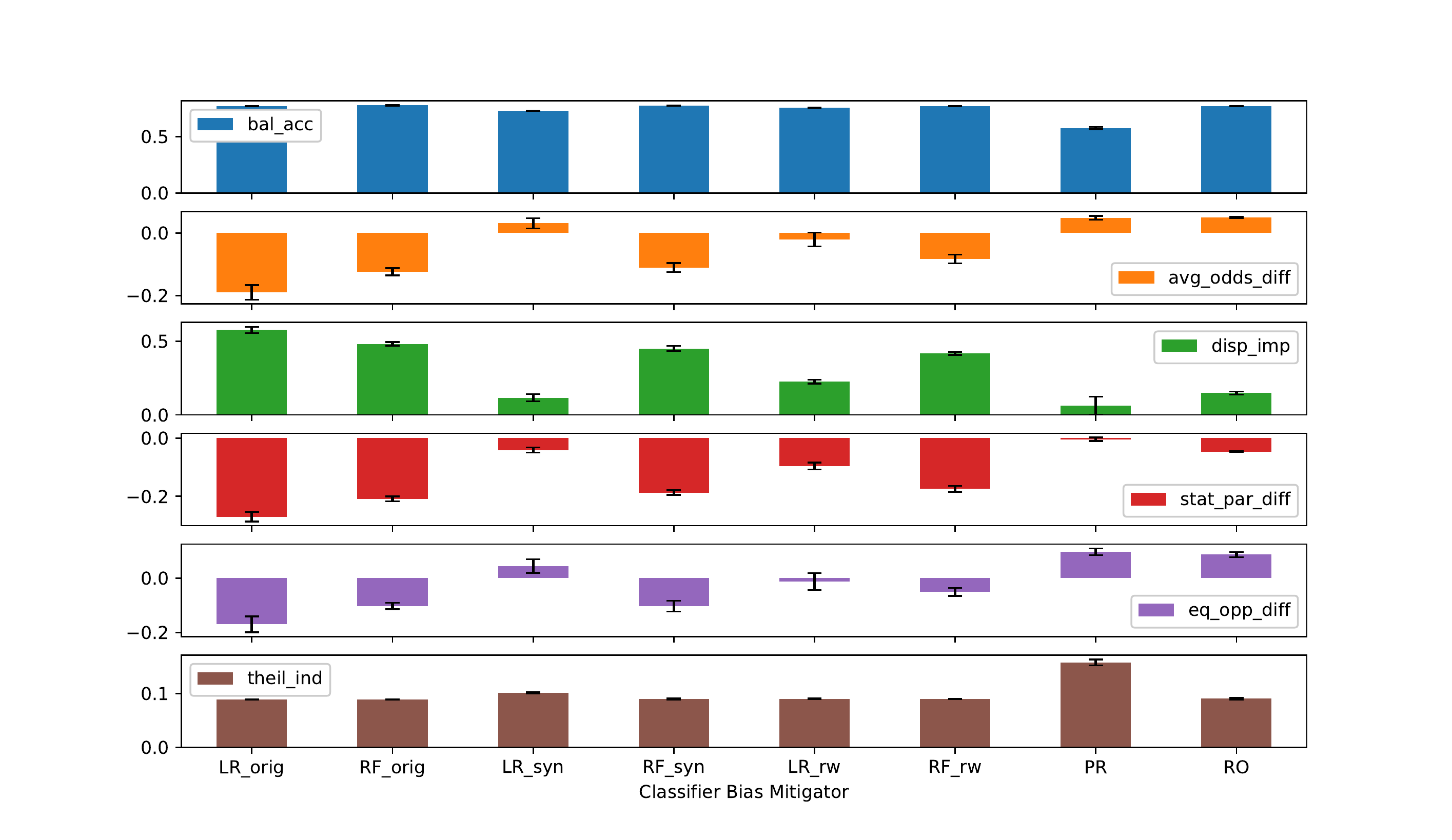}
\caption{\label{fig:mep_case2}Medical Expanse Price: Expand Privileged Unfavored.}
\end{subfigure}
\caption{\label{fig:mep} Fairness results on {\em Medical Expense Price} data.}
\end{figure}

\subsection{Bank Data}
\label{subsec:bank}

\begin{figure}[!htb]
\centering
\includegraphics[width=0.475\textwidth]{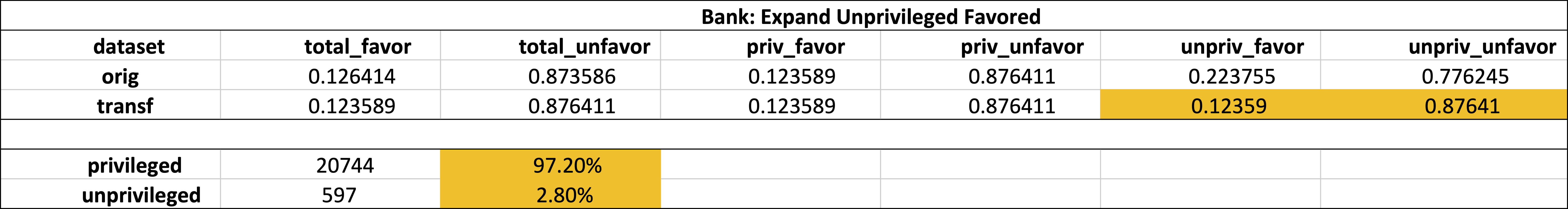}
\caption{\label{fig:bank_prior} Bank: favored/unfavored class distribution.}
\end{figure}

The favored/unfavored class distribution of the {\em Bank} data before and after oversampling is shown in Figure~\ref{fig:bank_prior}. The favored class is {\em subscribe deposit} and the privileged group is {\em `age $\ge$ 25'}. Notice the privileged group is overwhelmingly dominant in this dataset. 

%\clearpage

\begin{figure}[!htb]
\centering
\includegraphics[trim=2.5cm 0.5cm 2cm 1.5cm, clip, width=0.35\textwidth]{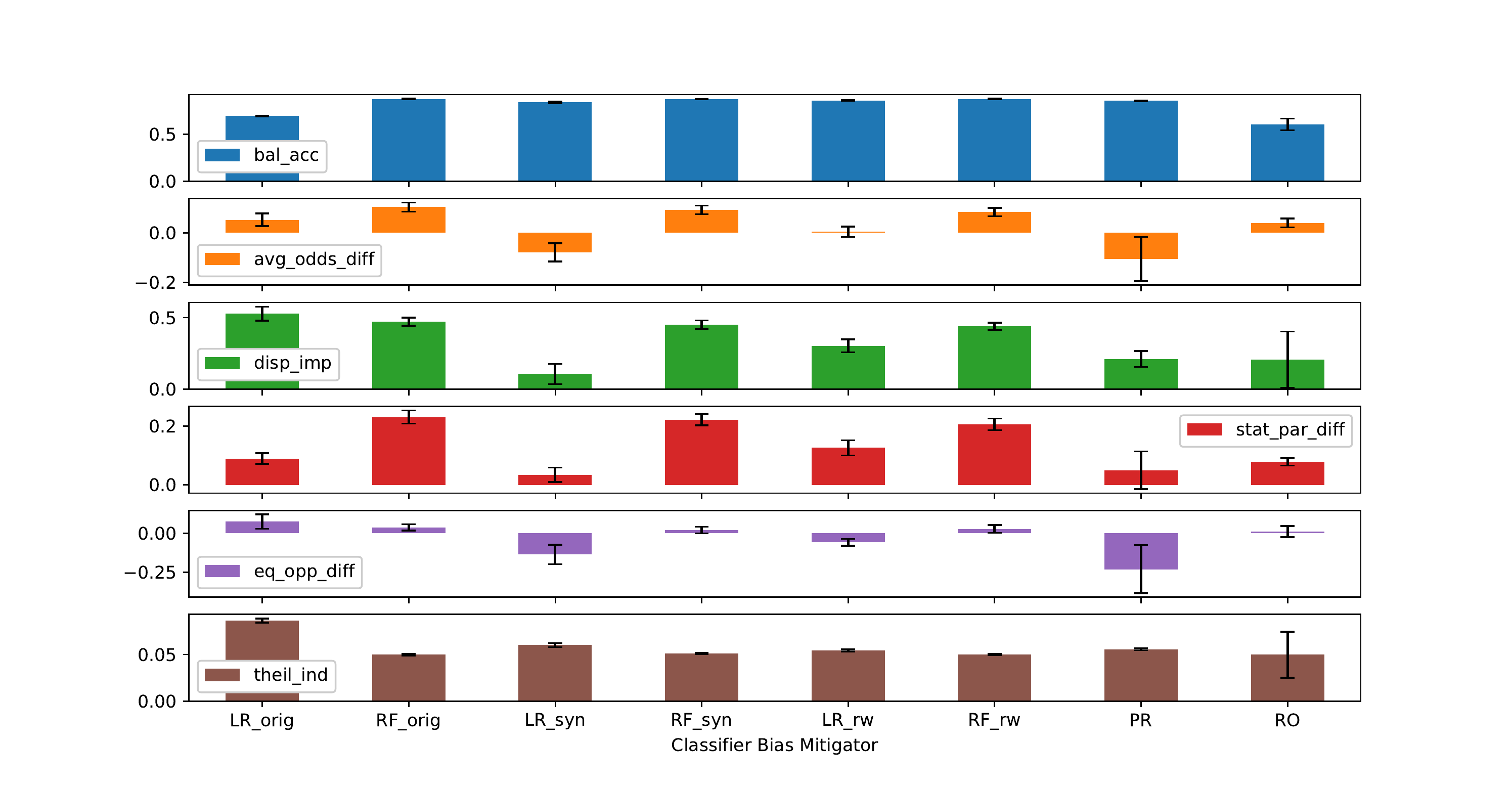}
\caption{\label{fig:bank} Fairness results on Bank data.}
\end{figure}

Figure~\ref{fig:bank} shows  our de-biasing results when we oversample the favored class for the underrepresented group. Our de-biasing technique is the most effective one in terms of overall performance. Our de-biasing technique clearly outperforms the {\em Prejudice Remover} (PR) technique. The {\em Reject Option} (RO) technique is comparable to ours in terms of fairness, however, at the price of much lowered accuracy and significantly larger variances. 
%\clearpage

\subsection{Additional Experiment}

\begin{figure}[!hbt]
    \centering
    \begin{minipage}{0.235\textwidth}
    \includegraphics[trim=40 20 40 40, clip, width=\textwidth]{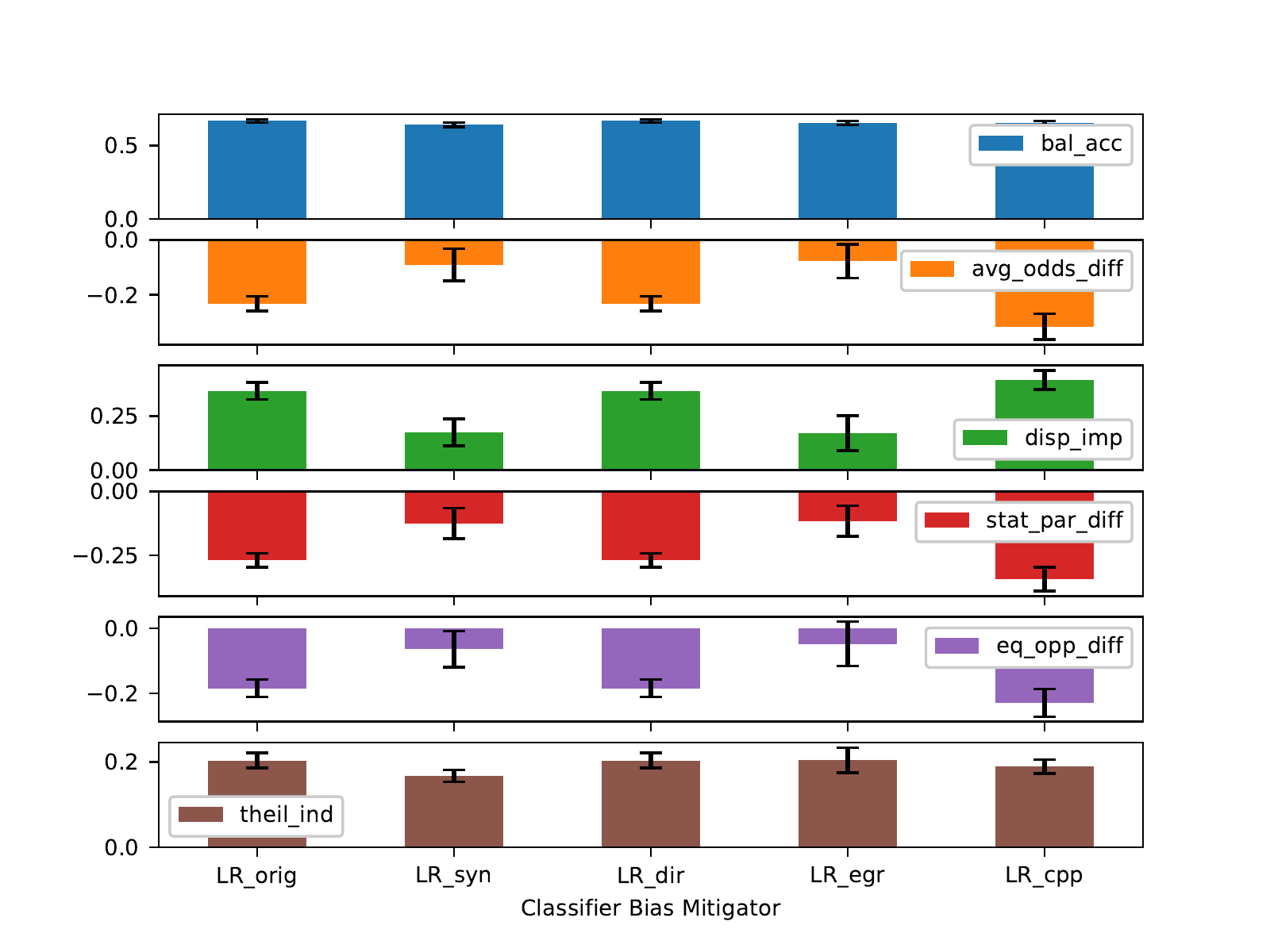}
    \vspace{-0.25in}
    \caption{LR ($p<0.0001$)}
    \label{fig:lr}
    \end{minipage}\hfill
    \begin{minipage}{0.235\textwidth}
    \includegraphics[trim=40 20 40 40, clip, width=\textwidth]{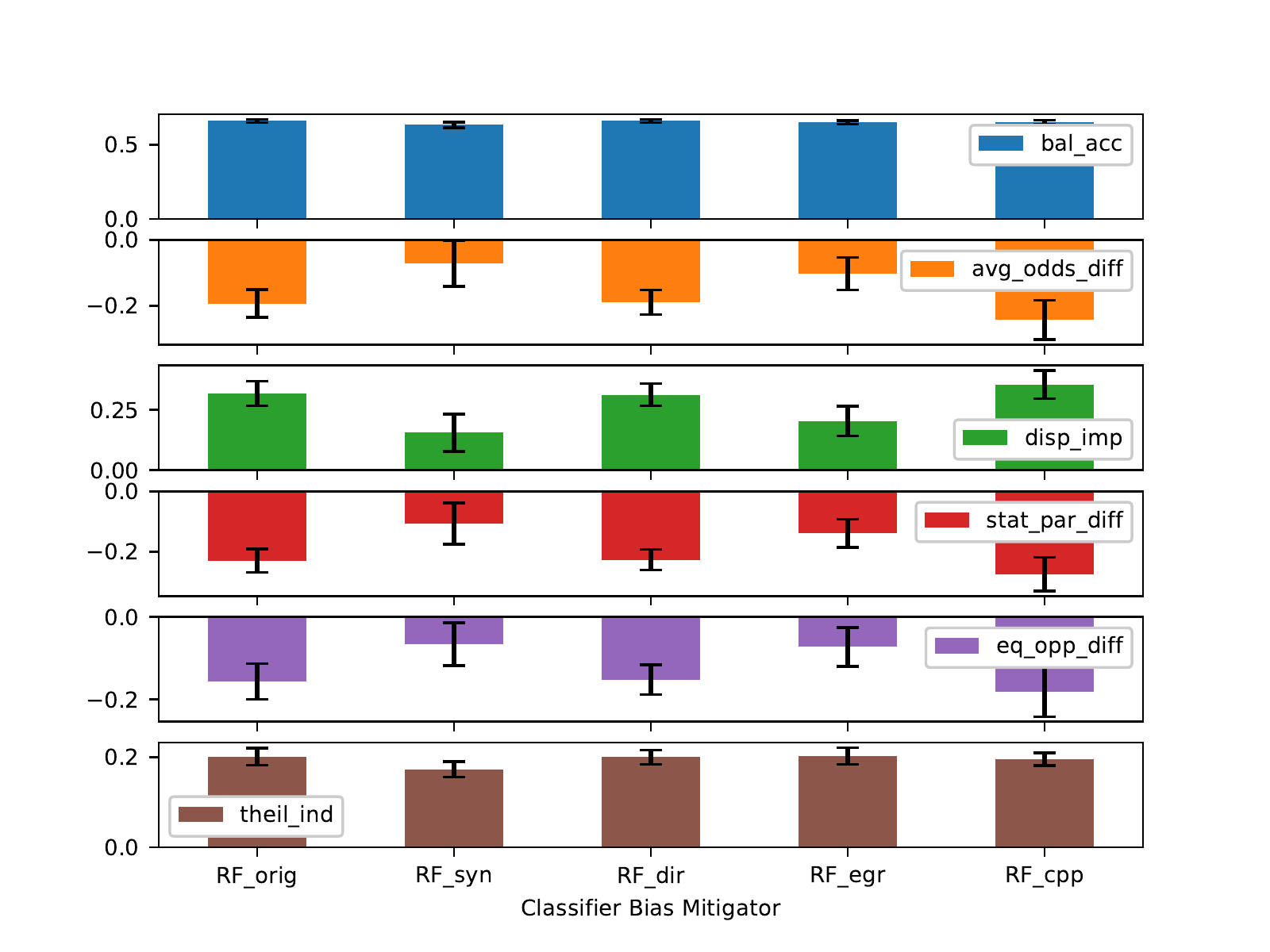}
    \vspace{-0.25in}
    \caption{RF ($p < 0.002$)}
    \label{fig:rf}
    \end{minipage}
    \begin{minipage}{0.235\textwidth}
    \includegraphics[trim=40 20 40 40, clip, width=\textwidth]{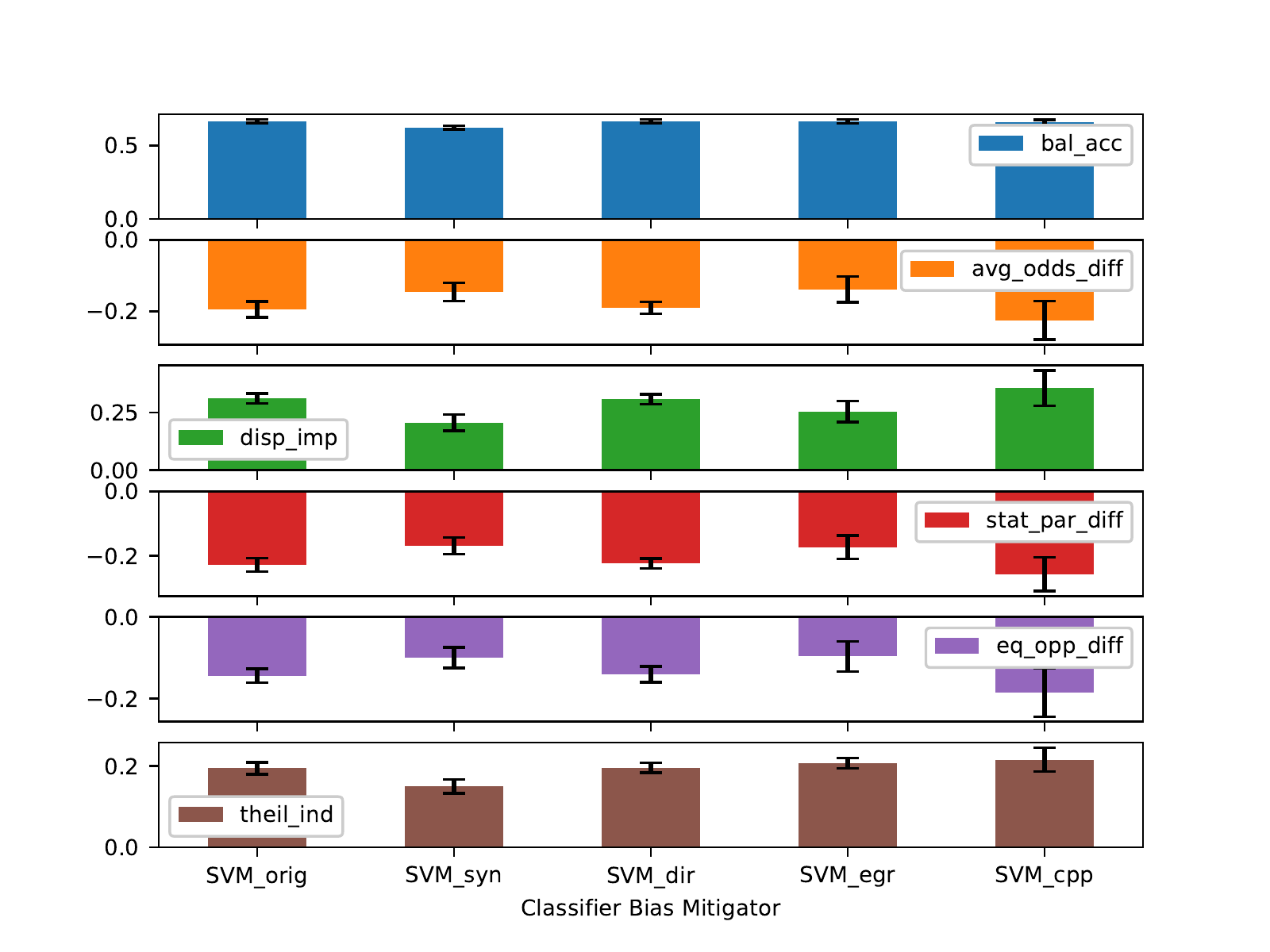}
    \vspace{-0.25in}
    \caption{SVM ($p < 0.001$)}
    \label{fig:svm}
    \end{minipage}\hfill
    \begin{minipage}{0.235\textwidth}
    \includegraphics[trim=40 20 40 40, clip, width=\textwidth]{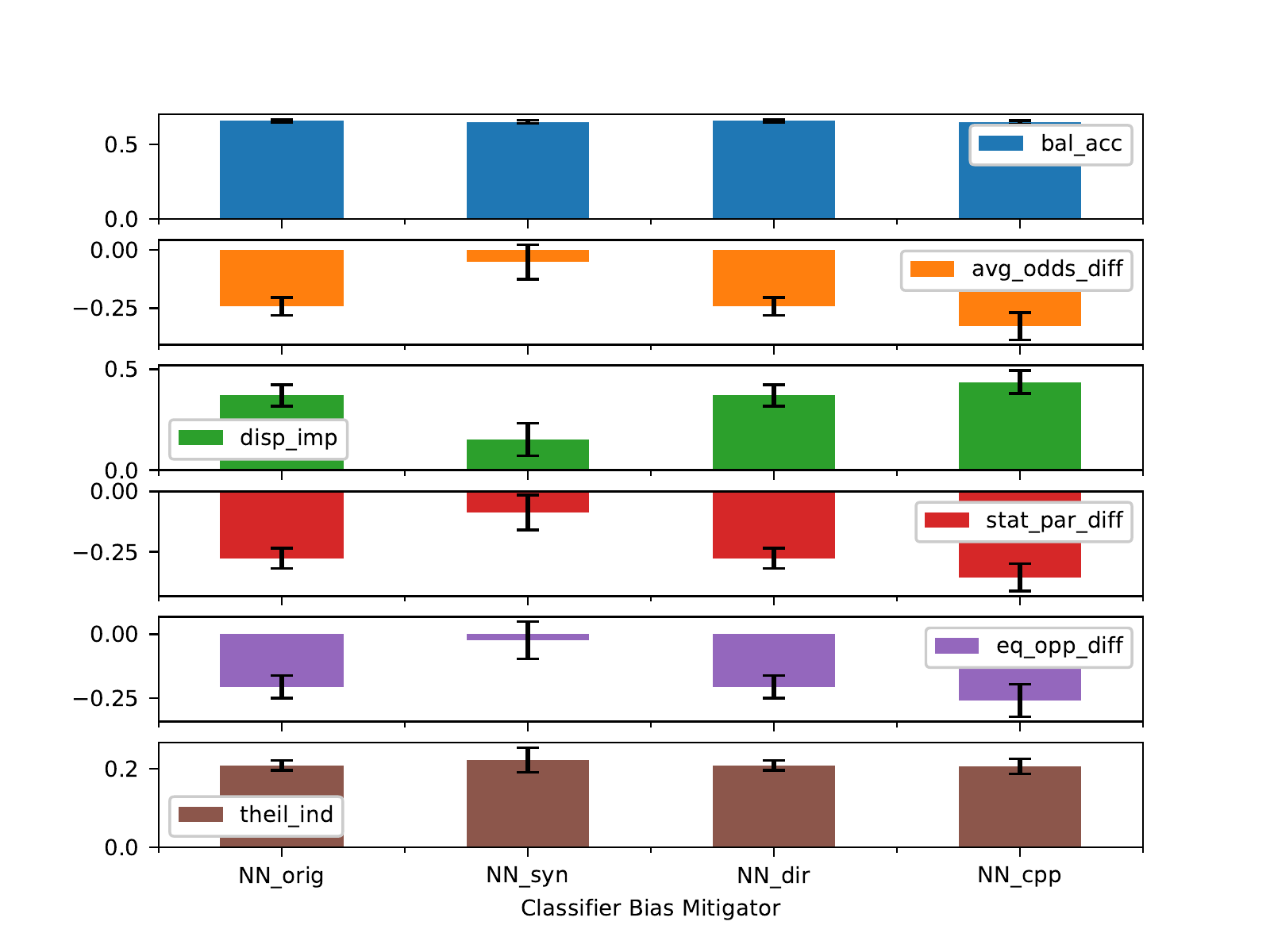}
    \vspace{-0.25in}
    \caption{NN ($p < 0.002$)}
    \label{fig:nn}
    \end{minipage}
\end{figure}

We also ran additional experiments on the {\em Compas} data to investigate the effectiveness of our technique when additional baselines SVM (Fig.~\ref{fig:svm}) \& Neural Net (NN) (Fig.~\ref{fig:nn}) and three other mitigators {\em  Disparate Impact Remover}  (pre-processing, denoted as \_dir (3rd col)), {\em Exponentiated Gradient Reduction} (in-processing, denoted as \_egr (4th col)), and {\em Calibrated EqOdds} (post-processing, denoted as \_cpp (5th col)) are added for comparison. Our technique (\_syn, 2nd col) consistently outperforms the others with paired t-tests $p < 0.002$ (except for Theil index $p \in [0.0001, 0.09]$) and accuracy drop $<$ 3\%. These results provide further strong evidence for the effectiveness of our approach.

\section{Conclusions and Future Work}
\label{sec:conclude}

Generating synthetic data to mitigate the lack of representation of the underrepresented group demonstrates a promising outcome for producing less biased learning models after training. Our approach focuses on expanding the training data in the area where positive base rate difference is originated, without violating the class priors. Compared to the existing mitigation techniques, our approach produces better overall fairness without a significant loss of accuracy. In the future, we plan to explore whether oversampling  can be used as a way of addressing structural biases,  for example, capturing the over-policing of minority areas, and creating synthetic data to compensate for the broader systemic biases.   

\bibliographystyle{named}
\bibliography{fair}

\end{document}